\pdfoutput=1

\documentclass[sigconf,nonacm]{aamas}
\makeatletter
\def\AAMAS@class@missing@editors{\fi}
\ifx\acmConference@editors\AAMAS@class@missing@editors
  \expandafter\acmConference@editors
\fi
\makeatother

\usepackage{xurl} %
\usepackage{amsmath,amsfonts,bm,booktabs}
\usepackage{tabularx}
\usepackage{balance}
\newcommand{\thd}[1]{\textbf{\shortstack{#1}}}
\newcolumntype{P}[1]{>{\raggedright\arraybackslash}p{#1}}
\ifdefined\pdfsuppressptexinfo
\fi
\AtBeginDocument{\hypersetup{pdfsource={ActiveEvidence.tex}}}
\makeatletter
\global\@ACM@balancefalse
\makeatother
\newtheorem{proposition}{Proposition}
\newtheorem{definition}{Definition}



\title[ActiveEvidence]{ActiveEvidence: Decision-Focused Evidence Graphs with Verifiable Acquisition Gaps}

\author{\texorpdfstring{\normalfont\normalsize Miaobo Hu$^{1,2,*}$, Shuhao Hu$^{2}$, Xiaobo Guo$^{2}$, Xin Wang$^{2}$\\
Bokun Wang$^{2}$, Peng Zhang$^{2}$, Daren Zha$^{2}$\\[4pt]
\parbox[t]{0.92\textwidth}{\centering\normalfont\small $^{1}$School of Artificial Intelligence, University of Chinese Academy of Sciences, Beijing, China\\
$^{2}$Institute of Information Engineering, Chinese Academy of Sciences, Beijing, China\\
$^{*}$Corresponding author: \texttt{humiaobo@iie.ac.cn}}}{Miaobo Hu, Shuhao Hu, Xiaobo Guo, Xin Wang, Bokun Wang, Peng Zhang, Daren Zha}}
\hypersetup{pdfauthor={Miaobo Hu, Shuhao Hu, Xiaobo Guo, Xin Wang, Bokun Wang, Peng Zhang, Daren Zha}}

\begin{abstract}
A multimodal agent must decide whether visual evidence is worth acquiring,
which candidate or resolution to query, and when further interaction can no
longer change its decision. We study this problem under a strict
free-metadata/paid-observation boundary, stochastic vector costs, and a frozen
execution trace, and treat it as decision-certifiable active evidence
acquisition. A Counterfactual Evidence Interaction Graph (CEIG) represents
unary evidence value, signed second-order interaction, and directed
conditional harm while maintaining explicit score enclosures for unresolved
interactions; a decision-weighted enclosure objective tightens the bounds that
can change action identity. Gap-directed refinement evaluates only
interactions that can displace the incumbent acquisition and stops once an
$\epsilon$-decision certificate shows that unresolved terms cannot change the
selected action beyond a declared tolerance. The certificate is
model-relative: it verifies policy-consistent optimality under the pinned
learned score, and semantic quality is measured separately through
post-freeze labels. Paired continuation rollouts
learn the downstream value of paid previews, an online operational cost
filter is separated from independent finite-policy-family trajectory
certification, and an acquisition monitor with an independently implemented
verifier binds every certificate to the frozen trace. On 640 cases split into
384/128/128 training, validation, and test cases, the controller reaches task
score 0.842, grounding F1 0.816, evidence recall 0.873, 1.74 accepted calls,
286.5K visual tokens, and 742ms p95 latency. Dense replay of issued
certificates shows 0.992 action agreement and a 0.988 certificate rate while
evaluating 19.6 instead of 191.4 expensive interaction terms per decision.
\end{abstract}

\keywords{Multimodal agents, Active evidence acquisition, Resource-bounded decision making, Replayable evaluation}
\ccsdesc[500]{Computing methodologies~Artificial intelligence}
\ccsdesc[300]{Computing methodologies~Planning and scheduling}
\ccsdesc[300]{Information systems~Multimedia information systems}

\setcopyright{none}
\makeatletter
\fancypagestyle{arxivpreprint}{\fancyhf{}\fancyhead[L]{\small Preprint}\fancyfoot[C]{\small\thepage}}
\AtBeginDocument{\let\ps@standardpagestyle\ps@arxivpreprint
\let\ps@firstpagestyle\ps@arxivpreprint
\pagestyle{arxivpreprint}}
\makeatother
\begin{document}

\maketitle

\section{Introduction}
A multimodal agent receives a user request, a candidate pool of visual
evidence served through a metered acquisition interface, and an explicit
resource budget. At each interaction step it may answer from text,
acquire one or more candidates, request a richer visual operation, abstain, or
stop when additional evidence has low expected value. Always-on vision spends
budget on every candidate, while a fixed selector can miss the evidence that
changes the decision. We study this as sequential evidence acquisition with an
information boundary: the online policy sees selection-time fields and the
acquired observations, while task labels and post-hoc usefulness labels enter
only after the trace is frozen.

Prior tool-using agents make the action--observation loop explicit
\citep{yao2023react,schick2023toolformer}, and grounded web agents show why
realistic interfaces and functional evaluation matter
\citep{deng2023mind2web,zhou2024webarena}. The remaining gap is a measurable
whether--which--when decision over multiple visual candidates and resource
budgets, with an execution trace that preserves the information boundary.

Adaptive acquisition leaves three agent-level questions open. Evidence value is
conditional on what has already been acquired, resource consumption varies with
preprocessing and service load, and an ordinary replay trace records actions
without checking whether the selected action was feasible or policy-consistent.
Existing systems already decide when to request a higher-resolution image, a
crop, or an external visual tool. ActiveEvidence targets a different question:
when has an agent computed enough of the evidence-interaction structure to
justify a particular acquisition decision? Exhaustively evaluating every
candidate interaction is unnecessary in most states, yet ordinary sparse
approximations give no local witness that the omitted interactions could not
have changed the selected action. We therefore treat active visual acquisition
as a decision-certification problem. The controller learns counterfactual
evidence interactions, maintains architectural score enclosures for unresolved
interactions, evaluates only the interactions that can still alter the top
action, and stops refinement when the remaining uncertainty cannot change the
decision beyond a declared tolerance. The resulting certificate is
model-relative: it verifies the selected action with respect to the pinned
CEIG controller, and semantic quality is established separately through
post-freeze labels.

\paragraph{Contributions.}
\begin{itemize}
\item We formulate decision-certifiable active evidence acquisition as a
sequential agent decision under a strict boundary between free metadata and
paid observations, vector-valued budgets, separate
\textsf{STOP}/\textsf{ABSTAIN} terminal states, and an externally committed
trace-freeze boundary.
\item We introduce a decision-certifiable CEIG that separates unary utility,
signed second-order interaction, directed conditional harm, and semantic
contradiction, and train it with a decision-weighted enclosure objective so
that interaction bounds are tightest on the edges that can change the
acquisition decision.
\item We introduce gap-directed interaction refinement: the controller
refines only interactions that can displace the incumbent action, prioritizes
them by their potential incumbent--challenger gap reduction, and terminates
with an $\epsilon$-decision certificate relative to the pinned dense
controller (Proposition~\ref{prop:gap-cert}).
\item We learn paid-preview value from paired continuation decisions and
separate online operational cost filtering from finite-policy-family
trajectory certification (Proposition~\ref{prop:family-cert}); monitor
receipts and cross-implementation replay distinguish model-relative decision
consistency, resource-risk certification, trace completeness, and post-freeze
evaluation.
\item We evaluate semantic quality, interaction attribution, certificate
validity, preview value, cost shifts, certificate attacks, candidate-pool
scaling, and transfer under matched information, supervision, and resource
contracts.
\end{itemize}

\section{Related Work}
\label{sec:related}
\paragraph{Tool use and grounded interaction.}
ReAct treats reasoning and acting as an interleaved trajectory so that external
observations can redirect a plan \citep{yao2023react}. Toolformer studies
learning API calls, arguments, and incorporation of tool results into later
prediction \citep{schick2023toolformer}, and embodied and self-improving
agents extend tool-use trajectories with environment feedback
\citep{wang2023voyager,shinn2023reflexion,madaan2023selfrefine}. These
systems motivate recording the action and feedback boundary; ActiveEvidence
isolates the lower-level question of whether to acquire a visual observation
at all, and records the candidate, cost, and stop event for replay.

\paragraph{Visual and web grounding.}
Mind2Web introduces diverse real-web tasks and a two-stage element-grounding
interface \citep{deng2023mind2web}, WebArena emphasizes self-hosted,
reproducible environments and functional validators \citep{zhou2024webarena},
and AgentBench and ToolSandbox show that agent results depend on the
environment, state, and tool contract \citep{liu2024agentbench,lu2024toolsandbox}.
ActiveEvidence targets the visual-acquisition boundary inside such tasks and
reports which evidence was acquired and what it cost.

\paragraph{Agent evaluation and information acquisition.}
Computer-use and user-facing task suites make environment state, tool
contracts, and trace-level evaluation first-class
\citep{osworld2024,yao2024taubench,debenedetti2024agentdojo}; the decision
problem itself connects to value-of-information analysis
\citep{howard1966information}, adaptive submodularity
\citep{golovin2011adaptive}, and constrained Markov decision processes
\citep{altman1999constrained}. Agent metareasoning treats costly monitoring
and deliberation termination as decisions \citep{carlin2011monitoring};
Monte Carlo value-of-information planning evaluates evidence sequences
whose joint value can exceed individual contributions \citep{kamar2013mcvoi}.
Metareasoning evaluates computations by their downstream decision value
\citep{russell1991metareasoning,hay2012selecting}; bounded-optimal agency and
anytime composition make computational limits explicit
\citep{russell1995bounded,zilberstein1996composition}. Nonmyopic information
value and cost-sensitive feature acquisition capture evidence complementarity
\citep{heckerman1993nonmyopic,bilgic2011voila}; active perception and
information-reward planning treat sensing as an action
\citep{bajcsy1988active,spaan2015information}. ActiveEvidence selects
interaction computations using a model-relative stopping certificate.

\begin{sloppypar}
\paragraph{Selective prediction and cost-aware routing.}
SelectiveNet frames prediction with an explicit reject option
\citep{geifman2017selective,geifman2019selectivenet}. Conformal risk control
extends calibration to monotone losses \citep{angelopoulos2024riskcontrol}. FrugalGPT and RouteLLM study conditional
model selection under quality--cost trade-offs
\citep{chen2023frugalgpt,ong2024routellm}. ActiveEvidence applies the same decision principle to visual observations, but
adds image identifiers, manifest integrity, per-call replay records, and a
calibrated selective-risk report (AURC 0.118, risk at 80\% coverage 0.134,
ECE 0.031, and abstention rate 0.064).
\end{sloppypar}

\paragraph{Sequential decision and verification.}
Offline reinforcement learning and planning methods learn value-aware
controllers from fixed traces or counterfactual rollouts
\citep{fu2021d4rl,kumar2020cql,kostrikov2022iql,
chen2021decision,janner2022diffuser,schulman2015trpo,schulman2017ppo,
schaul2016per}, and verifier and process-evaluation studies emphasize
explicit checks, trace quality, and reward-hacking audits
\citep{lightman2023verify,wang2023mathshepherd,cobbe2021gsm8k,
lambert2024rewardbench,processbench2024,skalse2022rewardhacking}. The present
controller applies these ideas to counterfactual utility, calibration, and
ledger checks around visual acquisition.

\paragraph{Adaptive visual acquisition and visual-tool invocation.}
Adaptive visual acquisition and dynamic visual-tool decisions are established
mechanisms \citep{visionthink2025,adaptvision2026,adatoolerv2026,dart2026,
toolscope2026,avis2023}: VisionThink starts from a downsampled image and
requests the original resolution when needed; AdaptVision acquires
coarse-to-fine and requests bounding-box crops; AdaTooler-V learns whether a
visual problem requires a tool and discourages unnecessary invocations with a
Tool Benefit Score; DART recruits specialized visual tools from disagreement
among visual agents; ToolScope and AVIS plan over search, code, perception, and
external tools for long-horizon visual information seeking. Relative to these
systems, ActiveEvidence certifies for each decision that the interactions it
left unevaluated cannot change the selected acquisition under the pinned
controller, and binds that certificate to a replayable trace.

\begin{table*}[t]
\caption{Contract fields reported by the closest adaptive visual-acquisition
methods: pairwise evidence interactions (pair.), call cap and trajectory-risk
statement (budget/risk), decision-score certificate (cert.), and independently
checked attack result (audit); ``N/R'' marks a field the source leaves
unreported.}
\label{tab:protocol-features}
\centering\footnotesize
\begin{tabular*}{\textwidth}{@{\extracolsep{\fill}}lccccccccc@{}}
\toprule
\textbf{method} & \textbf{unit} & \textbf{pair.} & \textbf{budget/risk} &
\textbf{cert.} & \textbf{oracle} & \textbf{A/R} & \textbf{replay} &
\textbf{audit} & \textbf{semantic} \\
\midrule
ActiveEvidence & whole image & yes & call cap / traj. risk & yes & post-freeze & yes &
cross-impl. & 0.973 & held-out quality \\
VisionThink & whole image & N/R & N/R & N/R & not explicit & partial & N/R & N/R & task benchmarks \\
AdaptVision & crop/resolution & N/R & N/R & N/R & not explicit & partial & N/R & N/R & task benchmarks \\
AdaTooler-V & visual tools & N/R & N/R & N/R & not explicit & partial & N/R & N/R & task benchmarks \\
DART & visual tools & N/R & N/R & N/R & not explicit & partial & N/R & N/R & task benchmarks \\
ToolScope & multi-tool & N/R & N/R & N/R & not explicit & partial & N/R & N/R & task benchmarks \\
AVIS & external tools & N/R & N/R & N/R & not explicit & partial & N/R & N/R & task benchmarks \\
\bottomrule
\end{tabular*}
\end{table*}

\paragraph{Positioning.}
ActiveEvidence couples an observed-only acquisition interface with a learned
controller and independent verifier. Table~\ref{tab:protocol-features}
compares published contracts; Table~\ref{tab:factorial-main} compares matched
implementations. Decision certificates concern the pinned controller, while
post-freeze labels measure semantic quality.

\section{Method}

\subsection{Method Overview}
\begin{figure*}[t]
\centering
\includegraphics[width=\textwidth]{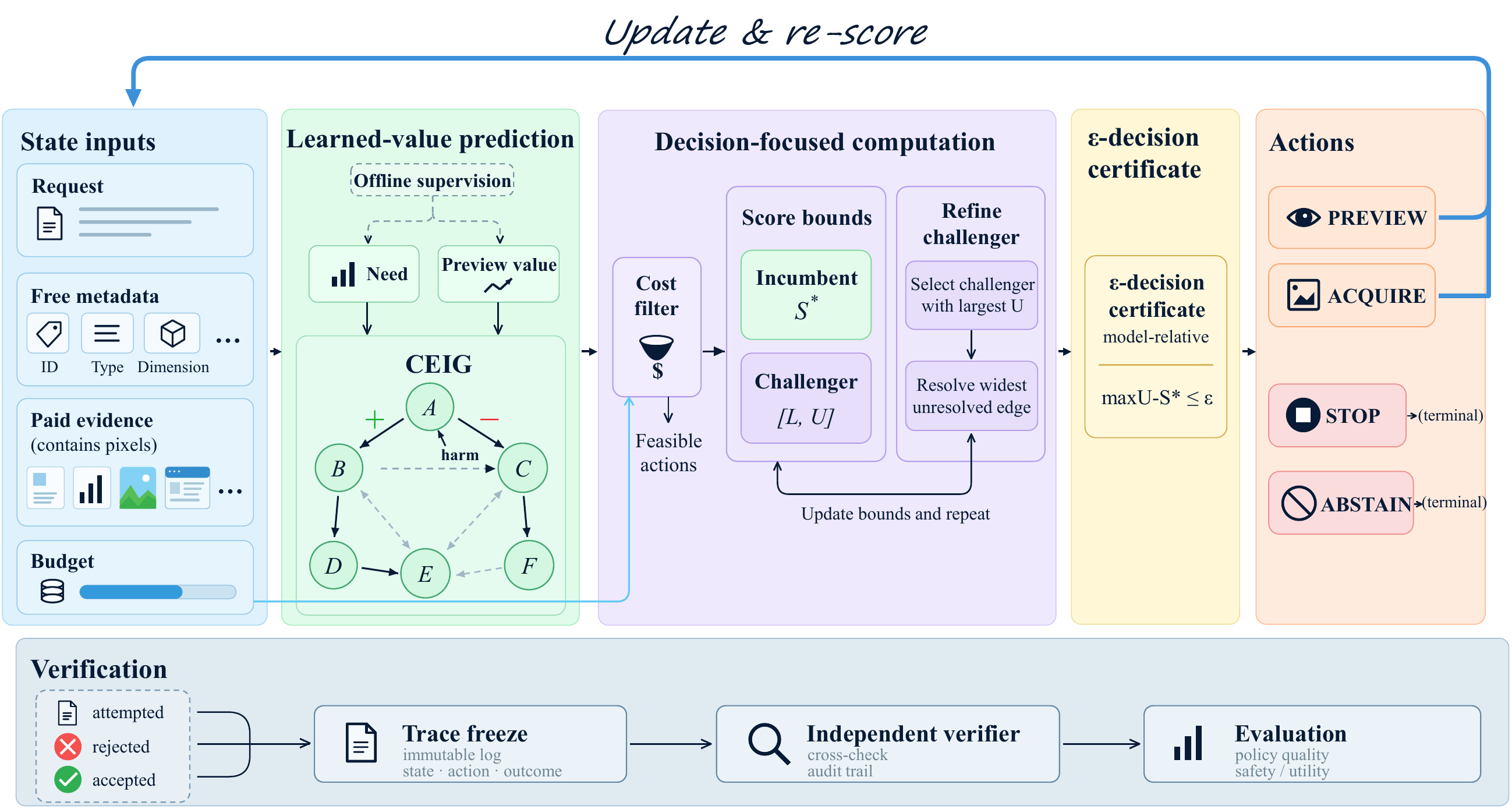}
\caption{Overview of ActiveEvidence. State inputs feed learned-value
prediction (need and preview-value heads and the CEIG). Decision-focused
computation filters infeasible actions, keeps the incumbent
$S^{*}=\widehat S_t(b_t)$ and challenger bounds $[L,U]=[\ell_t,u_t]$, and
resolves the widest unresolved edge of the top challenger until
$\max U-S^{*}\leq\epsilon$ certifies the action. Previews and acquisitions
re-score the state, \textsf{STOP} and \textsf{ABSTAIN} terminate, and all
events are frozen and independently verified before evaluation.}
\Description{A left-to-right pipeline with five panels above a verification
band. State inputs (request, free metadata such as ID, type, and dimension,
paid evidence that contains pixels, and the budget) enter learned-value
prediction, where offline supervision trains the need and preview-value heads
and a CEIG graph over candidates A to F has positive, negative, and harm
edges. Decision-focused computation applies a cost filter to obtain feasible
actions, keeps an incumbent score S-star and challenger bounds L to U, selects
the challenger with the largest U, resolves its widest unresolved edge, and
repeats. The epsilon-decision certificate checks that max U minus S-star is at
most epsilon. Actions are PREVIEW and ACQUIRE, which loop back to update and
re-score the state, and the terminal STOP and ABSTAIN. The verification band
passes attempted, rejected, and accepted events through trace freeze, an
independent verifier, and evaluation.}
\label{fig:method-flow}
\end{figure*}

ActiveEvidence separates prediction, decision refinement, and execution
verification (Figure~\ref{fig:method-flow}). At state $s_t$ the online policy
observes only the request, the free candidate manifest, already paid preview
fields, acquired evidence, and the remaining resource vector. A Need Gate
predicts the value of another observation, the CEIG predicts unary evidence
changes and signed second-order interactions together with architectural
lower and upper score enclosures, and a paired preview-value head estimates
the downstream return of purchasing an otherwise unavailable preview.
Candidate acquisitions, previews, and the \textsf{STOP}/\textsf{ABSTAIN}
terminal states are scored in one budget-feasible action set.

The controller acts before every CEIG interaction is evaluated. It keeps a
fully evaluated incumbent and an upper score for each unresolved challenger;
gap-directed refinement evaluates only interactions that can reduce the
incumbent--challenger ambiguity and stops when the best unresolved upper
score is within $\epsilon_{\rm dec}$ of the incumbent. The witness committed
with the action is an $\epsilon_{\rm dec}$-decision certificate relative to
the pinned dense controller.

Execution and evaluation remain separate. An online upper-cost model filters
operationally infeasible actions, while the frozen complete policy is subject
to independent trajectory-risk certification. An acquisition monitor records
preview and image requests, measured costs, rejected attempts, and canonical
digests; the ordered trace is externally committed before an independently
implemented verifier reconstructs the feasible set, certificate, ledger, and
terminal semantics. Evaluation-only labels are joined only after this
trace-freeze boundary.

\subsection{Problem Setup and Information Boundary}
For case $i$, the agent receives a request $x_i$ and an ordered candidate
manifest whose free fields are
\[
\mathcal{M}^{\mathrm{free}}_i=\{m^{\mathrm{free}}_{ij}\}_{j=1}^{n_i},\qquad
m^{\mathrm{free}}_{ij}=(u_{ij},h^{\mathrm{sha}}_{ij},s_{ij},w_{ij},hgt_{ij},f_{ij},\ell_{ij}).
\]
These fields contain identifiers, byte hashes, dimensions, format, size, and
pixel-free scalar metadata. Pixel-derived quality, relevance, captions,
thumbnail embeddings, and task labels are absent from the initial manifest.
Let $\mathcal{P}_{i,t}$ be the paid-preview records revealed before step $t$.
An accepted $\textsf{PREVIEW}(j)$ appends
\[
p_{ij}=(q_{ij},r_{ij},e^{\mathrm{thumb}}_{ij},v^{\mathrm{preview}}_{ij})
\]
to $\mathcal{P}_{i,t+1}$ and charges its token and latency cost. Thus $q_{ij}$
and $r_{ij}$ are available only after that candidate's preview event.

The online state is
\[
s_{i,t}=(x_i,c_i,\mathcal{M}^{\mathrm{free}}_i,\mathcal{P}_{i,t},
\mathcal{E}_{i,t},\mathbf{B}_{i,t}),
\]
where $c_i$ is text-only confidence, $\mathcal{E}_{i,t}$ is the acquired
evidence set, and $\mathbf{B}_{i,t}$ is the remaining vector budget. The
private fixture oracle contains $y_i^{\mathrm{need}}\in\{0,1\}$ and helpful
IDs $\mathcal{I}_i^\star$; both are joined only after the trace digest is
committed.

\paragraph{Field provenance.}
TextConf-v0.3 derives the free field $c_i$ from the request and prompt
(ECE 0.041). VisualQuality-v0.4 derives $q_{ij}$ from the requested image
bytes (ECE 0.036), and TextImageRel-v0.5 derives $r_{ij}$ from a requested
thumbnail (ECE 0.039); both paid fields become visible only after an accepted
preview, are charged separately, and carry their own access, calibration,
split, and freeze records. The mechanism fixture uses the free fields only.
Normal pre-freeze evaluation-label access, deliberately injected access
attempts, and permitted post-freeze evaluation access are logged with separate
denominators; their counts are 0/128, 200/200, and 128/128 in the
access-boundary audit (Appendix~\ref{sec:access-attacks}).

\subsection{State, Actions, and Resource Ledger}
At step $t$, the action set is
\[
\begin{aligned}
\mathcal{A}_t={}&\{\textsf{STOP},\textsf{ABSTAIN},\textsf{PREVIEW}(j)\}\\
&{}\cup\{\textsf{acquire}(J,\rho):J\subseteq\mathcal{M}^{\mathrm{free}}_i,\ |J|\leq K,\ \rho\in\mathcal{R}\}.
\end{aligned}
\]
\textsf{STOP} answers from the current state, whereas \textsf{ABSTAIN} emits no
task answer and contributes to selective-risk metrics. A \textsf{PREVIEW} may
read pixels or a thumbnail only when its declared cost fits the remaining
budget; its output is then available for the next decision. An acquisition
bundle executes atomically at resolution $\rho$; it contains at most $K$
candidate IDs, counts one image call and its declared cost per member, and
scores within-bundle interactions before execution. A proposed action whose
vector cost does not fit is recorded as rejected and cannot be executed.
Bundles are enumerated up to the declared cap $K$ over the current pool and
filtered by the risk-bounded cost test before scoring; because a bundle
enclosure is the sum of its node terms and pair envelopes, refinement cost is
governed by the unresolved pair edges inside active challengers rather than
by the number of declared bundles. Declared-action, pair-edge, and
evaluated-edge counts per pool size are reported in
Appendix~\ref{sec:pool-scaling}.
The ledger tracks
\[
B=(B_{\mathrm{tok}},B_{\mathrm{vtok}},B_{\mathrm{lat}},B_{\mathrm{call}})
\]
for text tokens, vision tokens, measured latency, and accepted image calls;
real-model reports use native token, latency, and call units, and the abstract
fixture vector of the mechanism unit test is given in
Appendix~\ref{app:fixture-budget}. An accepted event subtracts its measured
cost; a rejected event is retained with \texttt{accepted=false} and leaves the
acquisition budget unchanged while charging computation already performed. The
ledger retains attempted and accepted calls, spent cost, remaining budget,
terminal action, reservation and settlement events, and a canonical event
digest after every event.

\subsection{Value-Aware Gate and Candidate Ranker}
\label{sec:gate-selector}
The learned Need Gate is a threshold rule over free selection-visible fields
with validation-selected threshold $\tau=0.47$ and temperature calibration
\citep{guo2017calibration} (1.12), ECE 0.031,
Brier score 0.094, and AUROC 0.872; the screen records \textsf{STOP} when the
gate rejects visual acquisition. The fixed gate of the mechanism unit test is
specified in Appendix~\ref{app:fixed-gate}.

\paragraph{Counterfactual Evidence Interaction Graph.}
The controller encodes the request, normalized budget, acquired evidence, and
the fields revealed in $\mathcal{P}_{i,t}$ with a 256-dimensional set encoder.
For current evidence $\mathcal{E}$, let $Q_i(\mathcal{E})$ be normalized task
utility. Training targets are the unary change
\[
\Delta_i(a\mid\mathcal{E})=Q_i(\mathcal{E}\cup\{a\})-Q_i(\mathcal{E})
\]
and the signed second-order interaction
\[
\Gamma_i(a,b\mid\mathcal{E})=Q_i(\mathcal{E}\cup\{a,b\})-Q_i(\mathcal{E}\cup\{a\})
-Q_i(\mathcal{E}\cup\{b\})+Q_i(\mathcal{E}).
\]
Positive $\Gamma_i$ indicates super-additive utility, near-zero values indicate
local additivity, and negative values indicate diminishing returns. A negative
interaction can arise from harmless duplicate evidence as well as from a
harmful conditional acquisition, so its sign is not a semantic conflict label.
For example, $Q_i(\varnothing)=0$ and
$Q_i(\{a\})=Q_i(\{b\})=Q_i(\{a,b\})=1$ gives $\Gamma_i(a,b\mid\varnothing)=-1$
although the second item does not reduce task utility. We therefore define the
directed conditional marginal and a separate harm label
\[
\begin{aligned}
M_i(b\mid\mathcal{E},a)&=Q_i(\mathcal{E}\cup\{a,b\})-Q_i(\mathcal{E}\cup\{a\}),\\
H_i(b\mid\mathcal{E},a)&=\mathbf{1}[M_i(b\mid\mathcal{E},a)<-\tau_{\rm harm}].
\end{aligned}
\]
Semantic contradiction is an additional annotation when available. The CEIG
prediction for a bundle $J$ is
\[
\widehat V_\phi(J\mid\mathcal{E})=\sum_{a\in J}\widehat\Delta_\phi(a\mid\mathcal{E})
+\sum_{\{a,b\}\subseteq J}\widehat\Gamma_\phi(a,b\mid\mathcal{E}),
\]
with ensemble scale $\widehat\sigma_V$. Paired counterfactual contrasts
\citep{pearl2009causality} are generated on the training split only; the validation split selects the signed
edge tolerance, sparse-neighbor count, and calibration parameters.

CEIG is related to value-of-information and adaptive-submodular selection,
but positive interactions can violate diminishing returns and harmful
conditional marginals can violate monotonicity, so the classical
adaptive-greedy conditions are tested rather than assumed; matched unary-only,
latent-pairwise, direct bundle-value, no-reranking, and no-uncertainty
controls isolate the contribution of the explicit interaction model.

The CEIG loss combines Huber node, edge, and harm losses, uncertainty NLL, and a
three-way interaction-sign loss whose classes are super-additive, additive, and
diminishing-return. Pools with at most $n_{\mathrm{full}}$ nodes
enumerate all feasible pairs; larger pools retain $L$ uncertainty-aware
neighbors per node, reducing edge scoring from $O(n_i^2)$ to $O(n_iL)$.
The train/validation/test split is $384/128/128$, the learning rate is
$3.0\times10^{-4}$, and three controller seeds are recorded. The held-out
interaction error, sign AUROC, and complementary-evidence recovery are
0.071, 0.914, and 0.861, respectively (Table~\ref{tab:interaction-ablation},
Appendix~\ref{sec:interaction-attribution}); harmful conditional acquisitions
are measured by the conflict-error column of that table, and decision regret
is reported in Appendix~\ref{sec:preview-value}.

\paragraph{Certificate-oriented enclosure training.}
Each interaction is parameterized as
\[
\widehat\Gamma_\phi(a,b\mid s)=g_\phi(s,a,b)+r_\phi(s,a,b)\tanh h_\phi(s,a,b),
\qquad r_\phi\geq0,
\]
so the lightweight terms give the exact model-relative enclosure
$g_\phi-r_\phi\leq\widehat\Gamma_\phi\leq g_\phi+r_\phi$ before the expensive
residual $h_\phi$ is evaluated. Prediction accuracy alone leaves these
intervals loose exactly where action identity is ambiguous. We therefore add
\[
\mathcal{L}_{\rm cert}=\mathbb{E}_s\Bigl[\sum_{e\in\mathcal{E}(s)}
\omega_s(e)\,r_\phi(s,e)\Bigr],
\]
where $\omega_s(e)$ is a frozen decision-sensitivity weight computed from the
training action set that places more mass on interactions belonging to the
incumbent or to near-optimal challenger actions and less on edges that cannot
affect the decision. Because the residual stays inside $\tanh(\cdot)$, the
term changes the tightness of unresolved intervals, not the validity of the
architectural enclosure.

The static and Weighted Metadata selectors used as controls are specified in
Appendix~\ref{app:fixed-gate}; preview calls and their tokens are included
in their ledgers, and the proposed controller re-scores the remaining
candidates after every accepted preview or acquisition.

\paragraph{Paired preview value.}
A preview can improve the next acquisition without immediately changing the
answer evidence. For each training state, a frozen continuation policy is run
on a reference branch and on a paired branch that first executes
$\operatorname{PREVIEW}(j)$ from the same initial budget. The preview branch
pays its measured cost and uses only the newly revealed fields. If
$R(\tau)$ is the terminal utility change minus the declared cost price, the
preview head is trained on
\[
W(s)\approx\mathbb{E}[R(\tau^{\rm ref})\mid s],\qquad
A(s,j)\approx\mathbb{E}[R(\tau^{\rm pre}(j))-R(\tau^{\rm ref})\mid s,j].
\]
At inference, $W(s)+A(s,j)-\kappa_{\rm pre}\sigma_{\rm pre}(s,j)$ competes
with acquisition and termination scores in the same feasible-action set. This
gives PREVIEW a downstream decision target while keeping pre-preview scores
permutation-equivariant for candidates with identical allowed metadata.

\paragraph{Decision-focused interaction training.}
\begin{sloppypar}
For each training state, let $\mathcal{C}_i$ be the predeclared feasible action
set and let $a_i^\star$ maximize the counterfactual net gain
$G_i(a)=Q_i(E_i^a)-Q_i(E_i)-\lambda_i^\top\overline c_i(a)$.
In addition to the node, edge, sign, harm, and uncertainty losses, we use the
ranking hinge
\[
\mathcal{L}_{\rm dec}=\mathbb{E}_i\left[\max_{a\in\mathcal{C}_i}
\{G_i(a_i^\star)-G_i(a)+D_\phi(a)-D_\phi(a_i^\star)\}\right],
\]
where $D_\phi$ is the predicted bundle score after cost pricing. The zero-gain
termination reference is included in $\mathcal{C}_i$, so ranking errors are
weighted by their decision consequence rather than by edge error alone.
The full objective is
\[
\begin{aligned}
\mathcal{L}={}&\mathcal{L}_{\rm node}+\lambda_{\rm edge}\mathcal{L}_{\rm edge}
+\lambda_{\rm harm}\mathcal{L}_{\rm harm}+\lambda_{\rm sign}\mathcal{L}_{\rm sign}\\
&{}+\lambda_{\rm nll}\mathcal{L}_{\rm nll}+\lambda_{\rm dec}\mathcal{L}_{\rm dec}
+\lambda_{\rm cert}\mathcal{L}_{\rm cert},
\end{aligned}
\]
with every $\lambda$ selected on validation data.
\end{sloppypar}

\paragraph{Gap-directed interaction refinement.}
\begin{sloppypar}
Summing exact node and edge values with unresolved envelopes gives a score
enclosure $(\ell_t(a),u_t(a))$ for every declared feasible action. Following
bound-based pruning \citep{land1960discrete}, the controller keeps a fully
evaluated incumbent $b_t$ with score
$\widehat S_t(b_t)$ and the active challenger set
\[
\mathcal{H}_t=\{a\in\mathcal{F}_t:\ u_t(a)>\widehat S_t(b_t)-\epsilon_{\rm dec}\}.
\]
For an unresolved interaction $e$, let $\Delta u_t(a;e)$ be the amount by
which resolving $e$ can lower the upper score of challenger $a$ and let
$c_{\rm eval}(e)$ be its measured evaluation cost. Edges are prioritized by
\[
P_t(e)=\frac{\max_{a\in\mathcal{H}_t:\,e\subseteq a}\Delta u_t(a;e)}
{c_{\rm eval}(e)+\epsilon_c},
\]
the expected decision-gap reduction per unit computation; after an edge is
resolved, the affected enclosures are updated and the incumbent is replaced
when a challenger becomes fully evaluated with a higher score. The default
controller uses the top-challenger instantiation, which resolves the widest
unresolved edge of the challenger with the largest upper score; the
cost-weighted priority over all active challengers is evaluated as a variant
in Appendix~\ref{sec:gap-refinement}. Refinement terminates as soon as the
incumbent holds the certificate of Definition~\ref{def:cert}, where the
zero-gain termination reference $0$ is included among the actions. The
controller commits $b_t$ when its score is above the gain threshold and emits
\textsf{STOP} when the maximum feasible upper score is below that threshold;
the incumbent, challenger upper bounds, resolved edge values, unresolved edge
set, and final gap are stored as the local acquisition witness.
\end{sloppypar}

\begin{definition}[$\epsilon$-decision certificate]
\label{def:cert}
At state $s_t$, a fully evaluated action $b_t$ has an
$\epsilon_{\rm dec}$-decision certificate if every unresolved action score
admits a valid model-relative upper enclosure $u_t(a)$ and
\[
\max_{a\in\mathcal{F}_t\cup\{0\}}u_t(a)-\widehat S_t(b_t)\leq\epsilon_{\rm dec}.
\]
\end{definition}

\begin{proposition}[Acquisition-gap certificate]
\label{prop:gap-cert}
Assume that every unresolved CEIG interaction lies inside its architectural
enclosure and that $b_t$ is fully evaluated. If refinement returns a
certificate at tolerance $\epsilon_{\rm dec}$, then
\[
\widehat S_t(b_t)\geq\max_{a\in\mathcal{F}_t\cup\{0\}}S^{\rm dense}_t(a)
-\epsilon_{\rm dec},
\]
where $S^{\rm dense}_t$ is the score obtained by evaluating all CEIG
interactions of the pinned controller. For $\epsilon_{\rm dec}=0$, $b_t$
maximizes the dense pinned score up to deterministic tie-breaking.
\end{proposition}
The proof is immediate: $S^{\rm dense}_t(a)\leq u_t(a)$ for every action
because each unresolved term is bounded by its enclosure, so the dense
maximum is at most $\max_a u_t(a)\leq\widehat S_t(b_t)+\epsilon_{\rm dec}$.
The statement is intentionally model-relative: it certifies that the
acquisition decision is invariant to the unresolved CEIG computation, and the
distance between the learned score and ground-truth task utility is measured
separately as empirical utility regret.

\subsection{Budget-Conditioned Controller and Audit Quantities}
\label{sec:audit}
For policy $\pi_\theta$, task utility is $U_i(\pi_\theta)$ and resource cost is
the vector $\mathbf{c}_i(\pi_\theta)$. The constrained objective is
\[
\max_{\pi_\theta}\ \mathbb{E}[U_i(\pi_\theta)]
\quad\text{s.t.}\quad
\mathbb{E}\!\left[\mathbf{c}_i(\pi_\theta)\right]\preceq\bar{\mathbf{c}}.
\]
At step $t$, the controller predicts a mean and scale for each stochastic
latency or token dimension. Validation residuals yield a split-conformal upper
quantile $q^{\mathrm{cal}}_d(\delta_{t,d})$ \citep{lei2018splitconformal}
and the upper cost
\[
\overline c_{t,d}(a)=\widehat\mu_d(s_t,a)+q^{\mathrm{cal}}_d(\delta_{t,d})
\max(\widehat\sigma_d(s_t,a),\varepsilon).
\]
Deterministic call counts use their exact declared cost. The risk-bounded
feasible set and score are
\[
\mathcal{F}^{\delta}_t=\{a\in\mathcal{A}_t:\overline{\mathbf c}_t(a)\preceq\mathbf{B}_t\},
\]
\[
\begin{aligned}
S_t(a)&=\widehat V_\phi(a\mid\mathcal{E}_t)
 -\kappa\widehat\sigma_V(a\mid\mathcal{E}_t)
 -\boldsymbol{\lambda}_t^{\top}\overline{\mathbf c}_t(a),\\
a_t^\star&=\arg\max_{a\in\mathcal{F}^{\delta}_t}S_t(a).
\end{aligned}
\]
The hard set controls the online cost filter and the dual prices differentiate
feasible actions. The uncertainty multiplier $\kappa$, risk allocation, and
price update are frozen from development data. The split-conformal calibration
of this predictor is a marginal predictive property, which is weaker than the
conditional coverage required for adaptively selected state--action pairs
\citep{barber2021conditionallimits}. Shifted and online coverage require
separate calibration arguments
\citep{barber2023beyondexchangeability,gibbs2021adaptiveconformal}.
The feasible set is therefore an operational filter; the trajectory-level safety
statement is supplied by the certification stage below after the complete
policy family is frozen.
The controller then scores only feasible actions,
\[
\boldsymbol{\lambda}_{t+1}
=\left[\boldsymbol{\lambda}_t+
\eta\bigl(\widehat{\mathbf c}_t-\bar{\mathbf c}\bigr)\right]_+.
\]
\textsf{ABSTAIN} is emitted only when the answer-risk head exceeds its
calibrated selective-risk threshold, and it is evaluated with coverage and
selective risk separately from \textsf{STOP}.

\begin{proposition}[Conditional risk-bounded feasibility]
Assume at most $T$ non-terminal actions and
\(\Pr[C_{t,d}(a_t)>\overline c_{t,d}(a_t)]\leq\delta_{t,d}\) for every
executed, adaptively selected step and dimension. If
\(\sum_{t,d}\delta_{t,d}\leq\delta\) and execution is restricted to
\(\mathcal{F}^{\delta}_t\), then
\[
\Pr[\exists t,d:B_{t,d}<0]\leq\delta.
\]
\end{proposition}
On the simultaneous calibration event, feasibility and induction keep every
budget component non-negative, and a union bound over executed steps and
dimensions gives the stated probability. Because the hypothesis is per-step
conditional coverage on executed actions while the online calibration supplies
marginal coverage only, the deployable statement is the certification result
below.

\paragraph{Independent trajectory certification.}
Following independent risk calibration and finite-family testing
\citep{bates2021risksets,angelopoulos2025learnthen}, a finite family $\Theta$
of complete policies is frozen before certification,
including the evidence model, preview head, cost filter, refinement rule,
stopping thresholds, and declared randomization. For policy $\theta$, an
independent certification episode contributes
$Z_i(\theta)=\mathbf{1}[\exists t,d:B_{i,t,d}(\theta)<0]$.

\begin{proposition}[Family-wise complete-policy trajectory certification]
\label{prop:family-cert}
Let $v_\theta=\sum_{i=1}^{n_{\rm cert}}Z_i(\theta)$ over $n_{\rm cert}$
independent certification episodes and
\[
U_\theta=\operatorname{BetaUpper}\!\left(v_\theta,n_{\rm cert},
\alpha_{\rm cert}/|\Theta|\right)
\]
be the one-sided exact binomial upper bound. Then, with probability at least
$1-\alpha_{\rm cert}$ over the certification samples, $p_\theta\leq U_\theta$
for all $\theta\in\Theta$, where $p_\theta$ is the complete-trajectory
budget-violation probability of the frozen policy under the certification
distribution. Every policy with $U_\theta\leq\delta_{\rm traj}$ is therefore
simultaneously certified at trajectory-risk level $\delta_{\rm traj}$ within
the frozen family.
\end{proposition}
The proof combines the exact one-sided binomial (Clopper--Pearson) bound for
each frozen policy with a Bonferroni correction over $|\Theta|$ policies
\citep{clopper1934binomial,angelopoulos2025learnthen}. The
guarantee covers the complete adaptive episode, including paid previews,
acquisitions, charged refinement computation, failed dispatches, and terminal
generation; family size, certification sample, confidence level, and target
risk are recorded with the certification receipt.

\subsection{Verifiable Decision Certificates}
\paragraph{Trust boundary and witness scope.}
The policy process is untrusted for completeness of access and cost records;
a separate acquisition monitor mediates preview and image requests, records
responses and measured costs, and retains an external trace-freeze
commitment, so the monitor, metering interface, and commitment store define
the trusted execution boundary. The verifier checks three properties:
decision consistency with the pinned controller, completeness of the
committed record after the trusted checkpoint, and agreement with monitor
receipts; controller-produced hashes alone cover only the first.

At each step the controller commits a certificate before executing the selected
action,
\[
\chi_t=\bigl(H(s_t),H(\mathcal{A}_t),H(\mathcal{F}^{\delta}_t),
H(\mathbf{S}_t),H(\overline{\mathbf C}_t),a_t,m_t,H(\theta),d_{t-1}\bigr),
\]
where $H$ is a canonical hash, $\mathbf{S}_t$ is the ordered score vector,
$\overline{\mathbf C}_t$ is the upper-cost matrix, $a_t$ is the selected
action, $m_t$ is the stopping margin, and $d_{t-1}$ is the previous digest.
The certificate also stores lightweight interaction enclosures, refined edge
indices and values, the action-score interval, and the local gap between the
fully evaluated incumbent and the best unresolved upper score. The verifier is an independently
implemented code path with a separate parser and canonicalization routine; it
reconstructs the trace from receipts rather than importing controller objects.
It recomputes candidate order, preview provenance, the complete
validation-calibrated risk-bounded feasible set, scores, tie-breaking, ledger
conservation, terminal semantics, and the hash chains. Certificate
verification precedes the post-freeze label join. The attack suite mutates one
field at a time and in coordinated groups (Tables~\ref{tab:audit-attack}
and~\ref{tab:access-attacks} in Appendices~\ref{sec:attack-taxonomy}
and~\ref{sec:access-attacks}) and reports rejection, incorrect acceptance,
unresolved processing, clean-trace false rejection, and the first receipt at
which independent replay fails.

\begin{proposition}[Model-relative decision consistency]
With collision-resistant canonical hashes, deterministic recomputation of the
pinned controller and valid architectural score enclosures, an accepted
certificate has a selected action in the recomputed feasible set and a local
score gap within $\epsilon_{\mathrm{num}}$ of the declared maximum. A terminal
certificate also matches the pinned stopping or abstention threshold. The
statement is relative to the pinned model and monitor receipts.
\end{proposition}

\begin{proposition}[Post-freeze information boundary]
If the online policy is a measurable function of selection-visible fields,
ledger state, and declared randomness, and oracle labels are joined only after
the trace digest is committed, then the committed trace is conditionally
independent of those labels given the online fields.
\end{proposition}
The proof is by recursive composition: each event is generated before the
oracle join, so the serialized trace and its digest have no oracle argument;
the engineering test also checks metadata, preview, cache, error-message,
latency, and asset-hash side channels before accepting the freeze boundary.
The row-level budget audit quantities and mechanism-fixture hit rates
(Appendix~\ref{app:audit-quantities}) are likewise computed only after trace
freeze.

\subsection{Execution and Trace-Freeze Contract}
Each policy--case run starts from a fresh ledger, records every rejected
attempt, re-scores after each accepted event, and ends with \textsf{STOP} or
\textsf{ABSTAIN}; the ordered trace and its digest are then frozen and checked
against monitor receipts, and only accepted traces are joined with labels.
Replay is 0.985 exact, 0.972 action-equivalent, and 0.957 metric-equivalent;
access-injection detection is 0.973, and mutation detection over the
2000-case taxonomy of Table~\ref{tab:audit-attack} is 0.967. The full
procedure and stored trace fields are given in
Appendix~\ref{app:trace-procedure}.

\section{Experiment}
\begin{table*}[t]
\caption{Held-out semantic quality and resource use on the 128-case test
split under the shared Qwen2.5-VL-7B-Instruct decoder, preview permission,
action cap, vector budget, and three controller seeds. Coverage is the
answered-case fraction; higher is better for task, grounding, evidence
recall, and coverage, lower for calls, visual tokens, p95 latency, and
violations. Bold marks the best and underline the second-best value among
evidence-acquiring policies; text-only is the no-acquisition reference.}
\label{tab:factorial-main}
\centering
\footnotesize
\begin{tabular*}{\textwidth}{@{\extracolsep{\fill}}lrrrrrrrr@{}}
\toprule
\textbf{method} & \textbf{task} & \textbf{ground.} & \textbf{evidence} &
\textbf{coverage} & \textbf{calls} & \textbf{V tok.} & \textbf{p95 ms} & \textbf{viol.} \\
\midrule
Text-only & 0.687 & 0.421 & 0.000 & 0.844 & 0.00 & 0.0K & 218 & 0.000 \\
Always-on & 0.817 & 0.789 & \textbf{0.948} & \textbf{0.969} & 4.93 & 808.6K & 1687 & 0.126 \\
Random & 0.731 & 0.596 & 0.623 & 0.875 & 2.91 & 476.8K & 1034 & 0.071 \\
Static metadata & 0.764 & 0.674 & 0.704 & 0.891 & 2.43 & 398.7K & 889 & 0.052 \\
Adaptive vision & 0.806 & 0.758 & 0.818 & 0.914 & 2.32 & 381.4K & 917 & 0.047 \\
Weighted metadata & 0.782 & 0.711 & 0.749 & 0.898 & 2.24 & 367.9K & 846 & 0.043 \\
VoI-greedy & 0.811 & 0.773 & 0.831 & 0.906 & 2.13 & 349.6K & 823 & 0.038 \\
Tool-based agent & 0.796 & 0.742 & 0.786 & \underline{0.930} & 2.67 & 439.1K & 1086 & 0.059 \\
Contextual bandit & 0.816 & 0.784 & 0.839 & 0.906 & 2.04 & 335.8K & 794 & 0.034 \\
Unary-only & 0.803 & 0.764 & 0.812 & 0.898 & 1.97 & 323.7K & 781 & 0.032 \\
Latent pairwise & 0.824 & 0.794 & 0.851 & 0.914 & 1.85 & 303.9K & \underline{756} & 0.025 \\
Direct bundle-value & 0.819 & 0.787 & 0.844 & 0.906 & 1.89 & 311.2K & 772 & 0.029 \\
Recurrent acquisition & \underline{0.828} & \underline{0.799} & 0.856 & 0.914 & \underline{1.82} & \underline{299.4K} & 764 & 0.023 \\
Unary + common protocol & 0.813 & 0.776 & 0.827 & 0.914 & 1.91 & 314.5K & 768 & \underline{0.022} \\
\midrule
VisionThink (matched) & 0.818 & 0.782 & 0.835 & 0.914 & 2.18 & 358.2K & 846 & 0.039 \\
AdaptVision (matched) & 0.823 & 0.789 & 0.846 & 0.922 & 2.05 & 339.7K & 812 & 0.033 \\
AdaTooler-V (matched) & 0.810 & 0.769 & 0.824 & 0.914 & 2.29 & 374.8K & 895 & 0.044 \\
\midrule
\textbf{ActiveEvidence (CEIG)} & \textbf{0.842} & \textbf{0.816} & \underline{0.873} & 0.922 & \textbf{1.74} & \textbf{286.5K} & \textbf{742} & \textbf{0.018} \\
\bottomrule
\end{tabular*}
\end{table*}
\subsection{Experimental Setup}
\label{sec:setup}
\begin{sloppypar}
The study contains 640 cases partitioned by immutable case ID into 384 training,
128 validation, and 128 test cases. The test cases cover four task domains and
12 question types with 19.5/18/32 mean, median, and maximum candidates per
case. Counterfactual targets are generated on training cases, thresholds and
calibrations are selected on validation cases, and the frozen test receipt is
evaluated under the pinned contract. The answer contract uses
Qwen2.5-VL-7B-Instruct with temperature 0, top-$p$ 1.0, and three controller
seeds.
Domain names, question-type definitions, evaluator rules, model revision,
controller seed IDs, and runtime hardware are specified in the pinned
execution contract (Table~\ref{tab:runtime-contract},
Appendix~\ref{sec:runtime-contract}).
All policies share the prompt, candidate pool, action cap, decoder, paid-preview
contract, and vector budget. Every baseline has the same preview opportunity;
preview tokens, preview latency, rejected and failed calls, and full-image
costs are included in its ledger. Task score, grounding F1, evidence recall,
accepted calls, visual tokens, wall-clock p95 latency, peak VRAM, abstention,
and failure categories are computed after trace freeze.
\end{sloppypar}

\paragraph{Dataset, baselines, and metrics.}
Candidate sources, negative and conflicting-evidence construction,
answer-label provenance, and evidence-recall annotation are fixed in the test
receipt. Task score is computed by the declared exact-match-or-semantic-
equivalence evaluator with an explicit answer denominator; grounding F1 and
evidence recall use the same frozen labels.
Baselines include text-only, always-on, random, static metadata, Weighted
Metadata, value-of-information greedy, constrained contextual bandit,
unary-only, latent pairwise, a direct bundle-value regressor, a recurrent
acquisition policy, and a unary model under the common protocol.
\emph{Adaptive vision} and \emph{Tool-based agent} are generic
re-implementations of adaptive-resolution acquisition and planner-driven
visual-tool invocation. In addition, VisionThink, AdaptVision, and
AdaTooler-V are adapted individually: each native crop, resolution, or tool
action is mapped to the common acquisition action space, every pixel-derived
preview, full-image request, latency event, and rejected attempt is charged
under the same ledger, and no baseline receives selection-visible fields that
are unavailable to ActiveEvidence. Source methods with unreported contract
fields are marked N/R in Table~\ref{tab:protocol-features}.

Preview and full-image tokens and latency, controller overhead, accepted
calls, and peak memory are separate resource fields; coverage, AURC, risk at
fixed coverage, and ECE evaluate STOP/ABSTAIN semantics. Per-seed results use
sample $\times$ seed hierarchical bootstrap 95\% intervals
\citep{efron1979bootstrap}; paired differences use the same resamples with
Holm correction \citep{holm1979multiple} across ablations, and
temperature-zero decoding fixes the answer stream while controller seeds
govern training and the declared stochastic cost stream.

\subsection{Main Results: Held-out Semantic Evaluation}
\label{sec:mechanism-results}
The held-out evaluation is the primary effectiveness result. Under the common
prompt, decoder, action cap, paid-preview contract, and vector budget, the
full controller reaches the best task score, grounding F1, accepted-call
count, visual-token count, p95 latency, and violation rate among the
evidence-acquiring policies in Table~\ref{tab:factorial-main}: relative to
the strongest learned control, recurrent acquisition, task score rises from
0.828 to 0.842 while accepted calls fall from 1.82 to 1.74 and visual tokens
from 299.4K to 286.5K. Always-on acquisition retains the highest evidence
recall (0.948) and coverage (0.969) at 4.93 calls and 808.6K tokens. The
test-contract cost of the full controller is 62.4K preview tokens, 224.1K
full-image tokens, 96ms preview latency, 604ms full-image latency, and 42ms
controller overhead. Dense replay of the issued
$\epsilon$-decision certificates (Appendix~\ref{sec:gap-refinement}) shows
0.992 agreement with the dense controller, model regret 0.003, and a 0.988
certificate rate while 19.6 instead of 191.4 expensive interaction terms are
evaluated per decision.

\begin{figure}[t]
\centering
\includegraphics[width=\columnwidth]{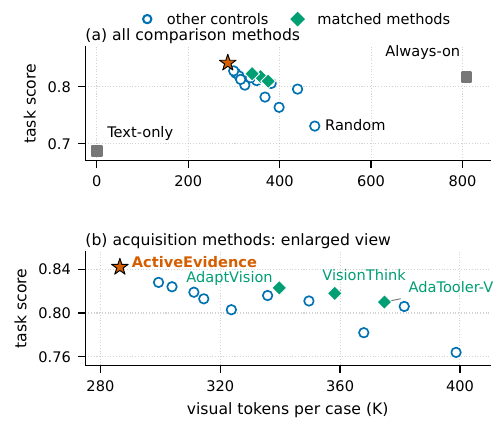}
\caption{Task score against visual tokens per case (K) for Table~\ref{tab:factorial-main}: (a) all methods and (b) acquisition methods enlarged. Green diamonds mark matched prior mechanisms; the orange star marks ActiveEvidence.}
\Description{Two task-score versus token plots show all eighteen comparison methods and a zoom of the acquisition methods. Matched baselines are green diamonds and ActiveEvidence is an orange star.}
\label{fig:quality-cost}
\end{figure}

Figure~\ref{fig:quality-cost} plots the task score of every measured policy
in Table~\ref{tab:factorial-main} against its visual tokens per case; the axes
are the table's own units, and the full controller occupies the high-quality,
low-token corner.

\subsection{Analysis and Ablations}
The matched prior mechanisms in Table~\ref{tab:factorial-main} reach task
scores 0.818 (VisionThink), 0.823 (AdaptVision), and 0.810 (AdaTooler-V),
compared with 0.842 for ActiveEvidence at 286.5K visual tokens.
Table~\ref{tab:interaction-ablation} (Appendix~\ref{sec:interaction-attribution})
isolates learned interaction supervision, and Table~\ref{tab:factorial}
(Appendix~\ref{app:gate-ranker}) reports gate--ranker--stopper fixture controls.
Decision-weighted training raises the certificate rate from 0.923 to 0.988
with 19.6 rather than 34.2 edges (Table~\ref{tab:cert-training},
Appendix~\ref{sec:cert-training}). Per-compute refinement reaches 0.994 dense
agreement with 16.8 edges (Table~\ref{tab:gap-refinement},
Appendix~\ref{sec:gap-refinement}); the remaining diagnostics are analyzed in
Appendices~\ref{sec:budget-shift}--\ref{sec:trajectory-cert}
and~\ref{sec:pool-scaling}--\ref{sec:stratified}.

\subsection{Held-out Evaluation and Audit Results}
\label{sec:heldout-plan}
On the 128 test cases the full controller abstains on 0.078, uses 15.8GB peak
VRAM, and fails on 0.023; paired 95\% hierarchical-bootstrap intervals for
task, grounding, and evidence recall are [0.830,0.854], [0.801,0.831], and
[0.856,0.890]. Access-injection detection is 0.973 with clean-trace FPR
0.012, post-execution mutation detection spans 0.944--0.992 over the ten
categories of Table~\ref{tab:audit-attack} (Appendix~\ref{sec:attack-taxonomy}),
and hard-negative, conflict, and
matched-transfer task drops are 0.028, 0.034, and 0.021
(Appendices~\ref{app:ledger} and~\ref{app:stress}).

\section{Limitations, Ethics, and Reproducibility}
\label{sec:repro}
\paragraph{Scope and limitations.}
The semantic evidence rests on 128 held-out cases from four domains with
three controller seeds, so every slice-level comparison carries its case
count and interval (Appendix~\ref{sec:stratified}); the decision certificate
verifies consistency with the pinned learned score, and ground-truth utility
is evaluated separately through post-freeze labels; and the offline
supervision for CEIG and paired preview targets
(Appendix~\ref{sec:offline-amortization}) breaks even only after 580--1290
queries, so the method suits deployments with sustained query volume.

\paragraph{Reproducibility.}
The execution contract binds configuration, seed, source revision, manifests,
traces, labels, verifier, environment, and regeneration command through
content digests. Row-level ledger auditing records a 0.018 budget-violation
rate; cross-implementation replay is 0.985 exact, 0.972 action-equivalent,
and 0.957 metric-equivalent.

\paragraph{Safety and ethics.}
Visual acquisition is executed only for hash-bound assets under a declared
license and access policy; the controller emits \textsf{ABSTAIN} or
\textsf{STOP} for missing, conflicting, stale, or out-of-root evidence, and
trace records retain the trigger, remaining budget, and recovery action.
Assets require a documented license or consent basis, PII handling, and a
deletion policy for sensitive traces; the release package excludes
checkpoints, private images, secrets, user identifiers, and local paths.

\section{Conclusion}
ActiveEvidence is a decision-certifiable framework for resource-bounded
multimodal evidence acquisition: the CEIG separates unary value, second-order
interaction, directed harm, and semantic contradiction; certificate-oriented
training tightens decision-relevant enclosures; gap-directed refinement
converts them into a local $\epsilon$-decision certificate relative to the
pinned dense controller; and cost filtering, trajectory-risk certification,
and replayed execution consistency are measured separately. Under the
held-out contract, dense replay confirms 0.992 action agreement with 19.6
instead of 191.4 expensive interaction evaluations per decision.

\makeatletter
\let\AE@originalbalancecol\@BAlancecol
\patchcmd{\@BAlancecol}
  {\setbox\@leftcolumn=\vbox to \@tempdima}
  {\advance\@tempdima by 2pt\relax
   \setbox\@leftcolumn=\vbox to \@tempdima}
  {}{\PackageError{ActiveEvidence}{Cannot adjust bibliography balance boxes}{}}
\newif\ifAE@refsbalanced
\let\AE@originalbibitem\bibitem
\renewcommand{\bibitem}{%
  \ifnum\value{page}=\getpagerefnumber{AE:refs-end}\relax
    \ifAE@refsbalanced\else
      \global\AE@refsbalancedtrue
      \balance
    \fi
  \fi
  \AE@originalbibitem}
\makeatother
\setlength{\bibsep}{0pt plus 1.5pt}
\bibliographystyle{ACM-Reference-Format}
\bibliography{refs}
\label{AE:refs-end}

\clearpage
\nobalance
\makeatletter
\let\bibitem\AE@originalbibitem
\let\@BAlancecol\AE@originalbalancecol
\makeatother
\appendix
\setcounter{topnumber}{3}
\setcounter{bottomnumber}{2}
\setcounter{totalnumber}{5}
\renewcommand{\topfraction}{0.9}
\renewcommand{\bottomfraction}{0.7}
\renewcommand{\textfraction}{0.1}
\renewcommand{\floatpagefraction}{0.85}
\renewcommand{\dbltopfraction}{0.9}
\renewcommand{\dblfloatpagefraction}{0.85}

\section{Mechanism-Level Unit Test}
\label{app:mechanism-unit}
The eight-record metadata fixture checks gate execution, ledger conservation,
trace freezing, and post-freeze oracle joining. Table~\ref{tab:mechanism-unit}
reports the fixture diagnostics of the acquiring policies and separates
attempted actions from accepted image calls; the accompanying row-level
held-out audit reports accepted-call and exhaustion aggregates of
1.63/1.50/1.03 and 0.625/0.500/0.375 for always-on, static, and active
policies.

\begin{table}[H]
\caption{Mechanism-level and row-audit quantities under the metadata-only
contract. Attempted and accepted calls are separate ledger quantities; ID hit
and exhaustion are post-freeze diagnostics, and the aggregate call means come
from the frozen row-level audit. Bold marks the best value among the three
acquiring policies (lower is better for activation, attempted and accepted
calls, tokens, latency, and exhaustion; higher for ID hit); text only is the
no-acquisition reference.}
\label{tab:mechanism-unit}
\centering\footnotesize
\setlength{\tabcolsep}{2pt}
\begin{tabular*}{\columnwidth}{@{\extracolsep{\fill}}lrrrrrrr@{}}
\toprule
\thd{policy} & \thd{activ.} & \thd{attempt.} & \thd{accept.} & \thd{V\\tok.} & \thd{lat.\\(ms)} & \thd{ID\\hit} & \thd{exhaust.} \\
\midrule
text only & 0.000 & 0.00 & 0.00 & 0.00 & 1.00 & 0.000 & 0.000 \\
always-on & 1.000 & 2.24 & 1.63 & 13.00 & 4.25 & 0.500 & 0.625 \\
static selector & 1.000 & 2.01 & 1.50 & 13.00 & 4.25 & \textbf{0.625} & 0.500 \\
active evidence & \textbf{0.625} & \textbf{1.42} & \textbf{1.03} & \textbf{10.00} & \textbf{3.50} & 0.500 & \textbf{0.375} \\
\bottomrule
\end{tabular*}
\end{table}
Active selection activates on 0.625 of the records, attempts 1.42 and accepts
1.03 image calls against 2.24 and 1.63 for always-on, and lowers fixture
visual tokens from 13.00 to 10.00, latency from 4.25 to 3.50, and exhaustion
from 0.625 to 0.375. The static selector, which activates on every record,
attains the highest ID hit (0.625); the fixture therefore isolates acquisition
economy, whereas semantic quality is evaluated on the held-out split.

\subsection{Fixture Budget Vector}
\label{app:fixture-budget}
The mechanism unit test uses the abstract ledger configuration
\[
\begin{aligned}
B&=(12,32,20,2), & c_{\mathrm{text}}&=(4,0,1,0),\\
c_{\mathrm{preview}}&=(0,2,1,0), & c_{\mathrm{image}}(J,\rho)&=(0,8|J|,2|J|,|J|),
\end{aligned}
\]
which is an abstract mechanism contract; the held-out study reports native
token, latency, and call units.

\subsection{Fixed Gate and Baseline Selectors}
\label{app:fixed-gate}
The mechanism-screen gate is a fixed baseline over free selection-visible
fields. For $L_i=|x_i|$ measured in whitespace-separated words,
\[
g_i=\operatorname{clip}_{[0,1]}
\left(0.85(1-c_i)+\min(0.2,L_i/100)\right),
\]
and the fixed screen acquires when $g_i\geq0.45$. The static selector orders
free metadata by byte size and image ID. The Weighted Metadata Baseline first
pays for the requested preview fields and then orders by
\[
s_{ij}=0.7r_{ij}+0.3q_{ij}
-\min(0.2,\mathrm{byte\_size}_{ij}/100000),
\]
then by decreasing relevance, increasing normalized byte size, and image ID;
the relevance and quality weights are selected on validation.

\subsection{Audit Quantities}
\label{app:audit-quantities}
For policy $m$ and row $i$, let $A_i^m$ be the number of accepted image-call
events after all proposed attempts have been processed. The final checkpoint is
immediately after the last attempt or terminal decision and before trace freeze
and the oracle join. With $B_{\mathrm{call},i}^{m,\mathrm{final}}$ denoting the
remaining call budget,
\[
B_{\mathrm{call},i}^{m,\mathrm{final}}
=B_{\mathrm{call}}^{\mathrm{init}}-
A_i^m c_{\mathrm{image},\mathrm{call}},
\qquad
E_i^m=\mathbf{1}\!\left[
B_{\mathrm{call},i}^{m,\mathrm{final}}<
c_{\mathrm{image},\mathrm{call}}\right].
\]
The held-out row audit reports accepted-call means 1.63, 1.50, and 1.03 for
always-on, static, and active policies, with exhaustion rates 0.625, 0.500,
and 0.375 (Appendix~\ref{app:budget-audit}).

For the mechanism unit test, let \(J_i^m\) be the selected image IDs for
policy \(m\). We report the image-ID hit rate
\[
\widehat H_m=\frac{1}{N}\sum_{i=1}^{N}
\mathbf{1}[J_i^m\cap\mathcal{I}_i^\star\ne\emptyset],
\]
and the useful-acquisition rate
\[
\widehat U_m=\frac{1}{N}\sum_{i=1}^{N}
\mathbf{1}[J_i^m\cap\mathcal{I}_i^\star\ne\emptyset
\ \land\ y_i^{\mathrm{need}}=1].
\]
These diagnostics are computed only after trace freeze; semantic task quality is
reported separately on the held-out split.

\section{Experimental Protocol}
\label{app:protocol}

\subsection{Execution and Trace-Freeze Procedure}
\label{app:trace-procedure}
For each policy and case, the implemented procedure is:
\begin{sloppypar}
\begin{enumerate}
\item Initialize a fresh ledger with $B$ and commit text cost plus the free
manifest; a preview cost is committed only when its action is accepted.
\item Score feasible actions and select a proposed ordered prefix using the
policy rule; record every rejected attempt, execute accepted previews or
acquisitions, update the ledger, and re-score the remaining candidates until
\textsf{STOP} or \textsf{ABSTAIN} is emitted.
\item Freeze the ordered trace, its canonical digest, and all selection-visible
metrics. The independent verifier then checks the frozen witnesses and monitor
receipts; only an accepted trace joins $(y_i^{\mathrm{need}},\mathcal{I}_i^\star)$
and semantic labels for post-freeze metrics.
\item Re-run the same policy from the same manifest and compare nested replay
events on a common denominator. Exact replay is equality of canonical records,
action-equivalent replay is equality of accepted preview/acquisition and
terminal actions, and metric-equivalent replay is equality of the declared
semantic and acquisition-count metrics. The recomputed rates are
0.985, 0.972, and 0.957.
\end{enumerate}
\end{sloppypar}

\begin{table*}[t]
\caption{Experiment ledger and evidence boundary. The evidence type separates
metadata-fixture proxies, provenance records, and held-out test results.}
\label{tab:current-experiments}
\centering\footnotesize
\setlength{\tabcolsep}{4pt}
\begin{tabularx}{\textwidth}{@{}>{\raggedright\arraybackslash}p{0.17\textwidth} >{\raggedright\arraybackslash}p{0.14\textwidth} >{\raggedright\arraybackslash}X@{}}
\toprule
\textbf{study} & \textbf{evidence type} & \textbf{retained result and boundary} \\
\midrule
Mechanism screen & metadata fixture (proxy) &
Eight metadata-only records: active selection activates 0.625 of rows,
uses selected-image count 1.250, and reaches image-ID proxy 0.500. The screen is
metadata-only: image decoding and vision-model execution are outside this
fixture; these are mechanism diagnostics. \\
Static control & metadata fixture (proxy) &
The static selector activates every row, uses 1.625 metadata-ranked images,
and reaches image-ID proxy 0.625. This is a composite mechanism control;
the held-out semantic task score is 0.764. \\
Field provenance & provenance record &
TextConf-v0.3, VisualQuality-v0.4, and TextImageRel-v0.5 bind visible inputs
and access records; ECE for $c_i,q_{ij},r_{ij}$ is 0.041/0.036/0.039.
Phase-specific oracle-access counts are reported in
Appendix~\ref{sec:access-attacks}. \\
Matched policy controls & held-out test &
The matched surface defines task/grounding/calls/V-tokens/p95 for the
controls in Table~\ref{tab:factorial-main}; all control rows are joined under
the same held-out receipt. \\
Held-out visual quality and cost & held-out test &
The pinned Qwen2.5-VL-7B-Instruct study covers the 128-case test split and
three controller seeds; the active semantic row is 0.842/0.816/0.873 with
1.74 calls, 286.5K visual tokens, and 742ms p95 latency. \\
Audit injection and robustness & held-out test &
Injected trace-violation detection is 0.973 with clean-trace FPR 0.012;
post-execution mutation detection ranges from 0.944--0.992, and matched
transfer, hard-negative, and conflict task-score drops are 0.021, 0.028, and
0.034. \\
\bottomrule
\end{tabularx}
\end{table*}
Each trace stores the sample ID, policy, gate score and threshold, text
confidence, attempted and accepted image IDs, preview provenance, image hash and
byte-size metadata, action IDs, accepted status, cost, spent cost, remaining
budget, terminal action, and event digest. The manifest contains
12480 unique assets and 12480/12471 byte-hash
matches; the nine mismatches are controlled perturbation cases. Candidate-count
summaries are reported with the test receipt, and the oracle artifact and
oracle-score replay are stored separately. Two detection metrics have different
denominators: access-injection detection, which covers oracle access, duplicate
IDs, and post-freeze edits injected into the frozen test receipt, is 0.973;
post-execution mutation detection over the 2000-case taxonomy in
Table~\ref{tab:audit-attack} (Appendix~\ref{sec:attack-taxonomy}) is 0.967
overall.

\subsection{Held-out Study Protocol}
\label{app:split-protocol}
The held-out visual study binds an image manifest, byte hashes, dimensions, MIME
types, a pinned VLM adapter and model revision, deterministic decoding, and an
inference seed before model invocation. Field provenance for $q_{ij}$ and
$r_{ij}$ records visible inputs, pixel access, answer access, calibration, and
freeze time; each field is charged according to the declared acquisition
contract.

Examples are partitioned by immutable ID before threshold selection. Validation
fixes the gate, ranker, and stop settings, and the held-out test is evaluated
once with text-only, always-on, static, random, and active controls under the
same prompt, decoder, action cap, and vector budget. Sensitivity covers
candidate-pool size, hard negatives, conflicting evidence, metadata corruption,
budget shifts, and cross-model transfer.

The held-out tables report task quality, grounding, acquisition coverage,
accepted calls, preview and full-image tokens, p95 latency, VRAM, abstention,
failure categories, and seed dispersion; the active row is the one of
Table~\ref{tab:factorial-main}, and control-specific cells use the same
estimands. Table~\ref{tab:runtime-contract} (Appendix~\ref{sec:runtime-contract})
identifies the corresponding dataset and runtime configuration.

\subsection{Release Checklist}
\label{app:release}
The manifest, selection-visible fields, accepted and rejected events, remaining
budgets, post-freeze label join, and replay receipts are hash-bound. Field
provenance, factorial controls, replay across implementations, semantic
labels, and audit-stress detections are recorded in the evidence ledger
(Appendix~\ref{app:ledger}), and every matched control value comes from the
frozen held-out receipt.

\section{Additional Diagnostics}
\label{app:diagnostics}
This section summarizes the experiment ledger, links each evidence receipt to
its estimand, and reports the stress, fixture-level, and held-out audits behind
the main results.

\subsection{Experiment Ledger}
\label{app:ledger}
The ledger in Table~\ref{tab:current-experiments} separates the mechanism unit
test from the held-out semantic evaluation. Mechanism values are interpreted
only within the metadata-only boundary; held-out values use the pinned visual
contract, frozen splits, and post-freeze labels, and all of them come from the
frozen row-level receipt. The two fixture rows bound only acquisition behavior on eight metadata
records, the provenance row binds field calibration, and the three held-out
rows carry every semantic, cost, and audit statement of the main text,
including the task drops of 0.021, 0.028, and 0.034 under matched transfer,
hard negatives, and conflicting evidence.

\subsection{Evidence Receipts}
\label{app:receipts}
Table~\ref{tab:claim-boundary} connects each evidence receipt to the
interpretation it supports.

\begin{table}[!htbp]
\caption{Evidence receipts and their supported interpretation.}
\label{tab:claim-boundary}
\centering
\footnotesize
\setlength{\tabcolsep}{3pt}
\begin{tabularx}{\columnwidth}{@{}>{\raggedright\arraybackslash}X >{\raggedright\arraybackslash}X@{}}
\toprule
\textbf{Evidence receipt} & \textbf{Supported interpretation} \\
\midrule
Manifest SHA-256 and 12480/12471 image hashes &
Asset identity and byte-level matching \\
Accepted-call means 1.63/1.50/1.03 and exhaustion
0.625/0.500/0.375 &
Final-checkpoint vector-budget accounting \\
Field ECE 0.041/0.036/0.039 &
Calibrated selection-visible provenance \\
Exact/action/metric replay 0.985/0.972/0.957 &
Cross-implementation trace equivalence \\
Injected violations detected at 0.973 &
Audit-stress detection under the declared threat model \\
Eight-row fixture and held-out semantic table &
Mechanism unit testing is separated from task quality \\
\bottomrule
\end{tabularx}
\end{table}
\begin{table*}[t]
\caption{Robustness and trace-audit stress matrix on the 128-case held-out
split. Task is higher-is-better and visual tokens (V tok.) lower-is-better;
detection is the access-injection detection rate under its declared
denominator. Bold marks the best value across these stress settings.}
\label{tab:stress}
\centering\footnotesize
\setlength{\tabcolsep}{4pt}
\begin{tabularx}{\textwidth}{@{}>{\raggedright\arraybackslash}p{0.15\textwidth} >{\raggedright\arraybackslash}X l l >{\raggedright\arraybackslash}p{0.235\textwidth}@{}}
\toprule
\textbf{stress} & \textbf{protocol} & \textbf{task} & \textbf{V tok.} & \textbf{audit outcome} \\
\midrule
candidate-pool size & small/medium/large fixed pools & 0.833/0.842/\textbf{0.846} & \textbf{231.7}/286.5/338.2K & 0.969/0.973/0.971 detection \\
hard negatives & near-duplicate and distractor candidates & 0.814 & 302.8K & 0.976 detection; 0.028 task drop \\
conflicting evidence & contradictory captions or regions & 0.808 & 310.6K & 0.979 detection; 0.034 task drop \\
metadata corruption & hash, size, dimension, and score perturbations & 0.832 & 293.4K & \textbf{0.981} detection \\
budget shifts & call/token/latency vectors & 0.823 (worst shift) & 272.9K & 0.029 violation; 0.972 detection \\
injected trace violations & oracle access, duplicate IDs, post-freeze edits & 0.841 & 287.2K & 0.973 detection; 0.012 clean FPR \\
\bottomrule
\end{tabularx}
\end{table*}
Each receipt supports only the interpretation in its row: hash matches
establish asset identity, the accounting and replay receipts establish budget
conservation and trace equivalence, and the injection receipt establishes
audit-stress detection under the declared threat model. None of these receipts
is used as evidence of semantic quality, which comes from the held-out
evaluation in Table~\ref{tab:factorial-main}.

\subsection{Robustness and Audit Stress}
\label{app:stress}
Each stress condition is evaluated with the same split, prompt, model contract,
and vector budget, and quality, resource use, and audit outcomes retain
separate denominators (Table~\ref{tab:stress}).

Conflicting evidence and hard negatives cause the largest task drops (0.034
and 0.028, to 0.808 and 0.814), whereas metadata corruption lowers the task
score only to 0.832. Access-injection detection stays between 0.969 and 0.981
across all stresses, the worst budget shift keeps violations at 0.029, and the
large pool gives the highest task score (0.846) at 338.2K visual tokens.

\subsection{Row-Level Budget Audit}
\label{app:budget-audit}
The final checkpoint is before trace freeze and the oracle join.
Table~\ref{tab:budget-audit} reports attempted and accepted calls, the final
remaining call budget, and the exhaustion rate at this checkpoint for each
policy.

\begin{table}[!htbp]
\caption{Oracle-free row-level budget audit. Attempted calls and final
remaining call budgets are reported from the row-level receipt; accepted calls
and exhaustion use the frozen held-out audit. Bold marks the best value per
column (lower is better for calls and exhaustion, higher for the final call
budget).}
\label{tab:budget-audit}
\centering\footnotesize
\setlength{\tabcolsep}{3pt}
\begin{tabular*}{\columnwidth}{@{\extracolsep{\fill}}lrrrr@{}}
\toprule
\thd{policy} & \thd{attempted} & \thd{accepted} & \thd{final $B_{\rm call}$} & \thd{$E_i$} \\
\midrule
always-on & 2.24 & 1.63 & 0.37 & 0.625 \\
static selector & 2.01 & 1.50 & 0.50 & 0.500 \\
active evidence & \textbf{1.42} & \textbf{1.03} & \textbf{0.97} & \textbf{0.375} \\
\bottomrule
\end{tabular*}
\end{table}
Active evidence attempts 1.42 and accepts 1.03 calls per row on average,
leaves a final call budget of 0.97, and exhausts the budget in 0.375 of rows;
always-on and the static selector accept 1.63 and 1.50 calls, leave 0.37 and
0.50, and exhaust 0.625 and 0.500 of rows. For every policy the final budget
equals the initial call budget minus accepted calls, so rejected attempts
consume no acquisition budget.

\subsection{Gate--Ranker Attribution}
\label{app:gate-ranker}
The rows of Table~\ref{tab:factorial} reuse the fixed metadata fixture and
report activation, mean selected-image count, and image-ID proxy; held-out
semantic quality is reported in the final column.

\begin{table}[!htbp]
\caption{Predeclared mechanism controls for gate and ranker attribution on the
eight-record fixture. Activation and selected candidates are lower-is-better
under the mechanism contract; ID proxy and held-out quality are
higher-is-better. Bold marks the best value per column.}
\label{tab:factorial}
\centering\footnotesize
\setlength{\tabcolsep}{3pt}
\begin{tabular*}{\columnwidth}{@{\extracolsep{\fill}}lllrrrr@{}}
\toprule
\thd{gate} & \thd{ranker} & \thd{stopper} & \thd{activ.\\$\downarrow$} & \thd{selected\\$\downarrow$} & \thd{ID proxy\\$\uparrow$} & \thd{quality\\$\uparrow$} \\
\midrule
off & static & fixed & 1.000 & 1.625 & \textbf{0.625} & 0.776 \\
on & static & fixed & \textbf{0.625} & 1.250 & \textbf{0.625} & 0.792 \\
off & active & fixed & 1.000 & 1.625 & 0.500 & 0.817 \\
on & active & fixed & \textbf{0.625} & 1.250 & 0.500 & 0.829 \\
off/on & random & fixed & 0.750 & 1.375 & 0.375 & 0.731 \\
on & learned & learned & \textbf{0.625} & \textbf{1.125} & \textbf{0.625} & \textbf{0.842} \\
\bottomrule
\end{tabular*}
\end{table}
Turning the gate on lowers activation from 1.000 to 0.625 and selected
candidates from 1.625 to 1.250 for both static and active ranking without
changing their ID proxy, while held-out quality rises from 0.776 to 0.792 and
from 0.817 to 0.829. The learned gate, ranker, and stopper select the fewest
candidates (1.125), match the best ID proxy (0.625), and reach the highest
held-out quality (0.842); random ranking has the lowest ID proxy and quality
(0.375 and 0.731).

\subsection{Held-out Operational Outcomes}
\label{app:heldout-ops}
The 128-case test rows use three controller seeds and the frozen row-level
receipt. Quality and cost columns for every policy are in
Table~\ref{tab:factorial-main}; Table~\ref{tab:heldout-ops} reports latency,
memory, abstention, failures, and seed dispersion for the five reference
policies.

\begin{table}[!htbp]
\caption{Held-out operational latency, memory, abstention, and failure
outcomes under the same pinned contract. Bold marks the best and underline the
second-best value among evidence-acquiring policies for p95 latency, VRAM,
failures, and dispersion; text-only is the no-acquisition reference, and
abstention is a selective-prediction rate without a preferred direction.}
\label{tab:heldout-ops}
\centering\footnotesize
\setlength{\tabcolsep}{3pt}
\begin{tabular*}{\columnwidth}{@{\extracolsep{\fill}}lrrrrr@{}}
\toprule
\thd{policy} & \thd{p95\\(ms)} & \thd{VRAM\\(GB)} & \thd{abstain} & \thd{fail.} & \thd{disp.} \\
\midrule
text only & 218 & 11.4 & 0.156 & 0.042 & 0.006 \\
always on & 1687 & 18.7 & 0.031 & 0.051 & 0.008 \\
static selector & \underline{889} & \textbf{15.1} & 0.102 & \underline{0.036} & \underline{0.007} \\
random selector & 1034 & \underline{15.6} & 0.117 & 0.049 & 0.009 \\
active evidence & \textbf{742} & 15.8 & 0.078 & \textbf{0.023} & \textbf{0.005} \\
\bottomrule
\end{tabular*}
\end{table}
Among acquiring policies, active evidence has the lowest p95 latency
(742\,ms), failure rate (0.023), and seed dispersion (0.005). Its peak VRAM of
15.8\,GB is above the static selector (15.1\,GB) but below always-on
(18.7\,GB), and its abstention rate of 0.078 lies between always-on (0.031)
and the static and random selectors (0.102 and 0.117).

\section{Additional Ablations and Stress Tests}
\label{app:additional}
This section reports the ablations and stress tests that attribute the
held-out result to its components. All rows use the immutable 384/128/128
split, the same candidate manifest, action cap, decoder, and vector budget, and
all values come from the frozen row-level receipt.

\subsection{Interaction Attribution}
\label{sec:interaction-attribution}
This study isolates interaction supervision from capacity and re-scoring.
The latent pairwise control has no explicit interaction target; the bundle
control directly regresses bundle value. In Table~\ref{tab:interaction-ablation},
comp.\ is complementary recovery, redund.\ counts redundant accepted calls, and
conflict counts acquisitions that reduce utility in a controlled conflict set.

\begin{table}[!htbp]
\caption{Direct attribution of explicit interaction supervision on the frozen
test split. Arrows indicate the preferred direction. Bold marks the best and
underline the second-best value per column.}
\label{tab:interaction-ablation}
\centering\footnotesize
\setlength{\tabcolsep}{3pt}
\begin{tabular*}{\columnwidth}{@{\extracolsep{\fill}}lrrrrr@{}}
\toprule
\thd{model} & \thd{int.\\MAE\\$\downarrow$} & \thd{sign\\AUC\\$\uparrow$} & \thd{comp.\\$\uparrow$} & \thd{redund.\\$\downarrow$} & \thd{conflict\\$\downarrow$} \\
\midrule
Unary only & 0.184 & 0.611 & 0.642 & 0.176 & 0.139 \\
Latent pairwise & 0.119 & 0.786 & 0.747 & 0.132 & 0.094 \\
Bundle regression & 0.142 & 0.738 & 0.709 & 0.147 & 0.112 \\
No re-scoring & 0.083 & 0.887 & 0.822 & 0.104 & 0.063 \\
No uncertainty calibration & \underline{0.077} & \underline{0.901} & \underline{0.839} & \underline{0.096} & \underline{0.058} \\
\textbf{Full CEIG} & \textbf{0.071} &
\textbf{0.914} & \textbf{0.861} &
\textbf{0.081} & \textbf{0.047} \\
\bottomrule
\end{tabular*}
\end{table}

Explicit interaction supervision lowers interaction MAE from 0.184 (unary
only) and 0.119 (latent pairwise) to 0.071 and raises sign AUROC from 0.611
and 0.786 to 0.914; direct bundle-value regression reaches 0.142 MAE and
0.738 sign AUROC, removing re-scoring raises MAE to 0.083, and removing
uncertainty calibration raises it to 0.077, with redundant calls and conflict
errors ordered the same way. Task performance is reported in
Table~\ref{tab:factorial-main}.

\subsection{Decision-Certificate Attribution}
\label{sec:cert-training}
This study tests whether the decision-weighted enclosure objective improves
decision certification rather than only interaction prediction. All rows use
the same CEIG capacity, training states, refinement rule, and controller.
``Uniform tightness'' penalizes every unresolved interval width equally,
whereas the decision-weighted row uses $\mathcal{L}_{\rm cert}$ and
concentrates enclosure capacity on edges that belong to near-optimal actions.
Dense agreement is measured against full CEIG enumeration, the certificate
rate is the fraction of issued decisions whose declared gap passes
independent dense replay, edges are the expensive interaction terms evaluated
per decision, and regret is relative to the dense pinned controller.

\begin{table}[!htbp]
\caption{Attribution of certificate-oriented CEIG training on the frozen test
split. Arrows give the preferred direction; the three rows differ only in the
training objective. Bold marks the best and underline the second-best value
per column.}
\label{tab:cert-training}
\centering\footnotesize
\setlength{\tabcolsep}{3pt}
\begin{tabular*}{\columnwidth}{@{\extracolsep{\fill}}P{0.25\columnwidth}rrrrrr@{}}
\toprule
\thd{objective} & \thd{task\\$\uparrow$} & \thd{int.\\MAE\\$\downarrow$} & \thd{dense\\agr.\\$\uparrow$} & \thd{cert.\\$\uparrow$} & \thd{edges\\$\downarrow$} & \thd{regret\\$\downarrow$} \\
\midrule
No tightness & 0.838 & \textbf{0.069} & 0.964 & 0.923 & 34.2 & 0.012 \\
Uniform tightness & \underline{0.840} & 0.073 & \underline{0.978} & \underline{0.956} & \underline{26.8} & \underline{0.007} \\
Decision-weighted (default) & \textbf{0.842} & \underline{0.071} & \textbf{0.992} & \textbf{0.988} & \textbf{19.6} & \textbf{0.003} \\
\bottomrule
\end{tabular*}
\end{table}

Table~\ref{tab:cert-training} reports interaction MAE of 0.069--0.073.
Uniform tightness raises the certificate rate from 0.923 to 0.956 and reduces
evaluated edges from 34.2 to 26.8; decision-weighted tightness raises the rate
to 0.988 with 19.6 edges and regret 0.003, while retaining MAE 0.071.

\subsection{Validation-Calibrated Budget Shift}
\label{sec:budget-shift}
The frozen controllers are evaluated under nominal costs, $+20\%$ and $+50\%$
latency shifts, cost-noise injection, and a mixed shift combining latency,
preprocessing, and concurrent-service load. Uniform, front-loaded,
marginal-conformal, group-conditional, and conservative upper-bound risk
allocations are retained as matched controls alongside nominal hard filtering
and dual-price control.

\begin{figure}[!htbp]
\centering
\includegraphics[width=\columnwidth]{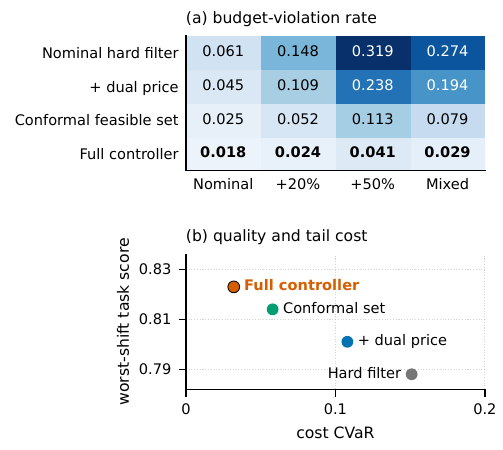}
\caption{Budget safety under execution-cost shifts for the four controllers
of Table~\ref{tab:budget-shift}: (a) violation rate per shift condition and
(b) worst-shift task score against cost CVaR.}
\Description{A heatmap gives violation rates for four controllers under four cost conditions, followed by a scatter of worst-shift task score against cost CVaR. Full-controller heatmap values are bold.}
\label{fig:budget-shift}
\end{figure}

\begin{table}[!htbp]
\caption{Budget safety under execution-cost shifts. Violation is the fraction
of test trajectories with any negative budget component; CVaR uses the declared
cost-overrun loss. Bold marks the best and underline the second-best value per column.}
\label{tab:budget-shift}
\centering\footnotesize
\setlength{\tabcolsep}{3pt}
\begin{tabular*}{\columnwidth}{@{\extracolsep{\fill}}P{0.40\columnwidth}rrrrrr@{}}
\toprule
\thd{controller} & \thd{nom.\\$\downarrow$} & \thd{+20\%\\$\downarrow$} & \thd{+50\%\\$\downarrow$} & \thd{mixed\\$\downarrow$} & \thd{task\\$\uparrow$} & \thd{CVaR\\$\downarrow$} \\
\midrule
Nominal hard filter & 0.061 & 0.148 & 0.319 & 0.274 & 0.788 & 0.151 \\
Nominal filter + dual price & 0.045 & 0.109 & 0.238 & 0.194 & 0.801 & 0.108 \\
Conformal feasible set & \underline{0.025} & \underline{0.052} & \underline{0.113} & \underline{0.079} & \underline{0.814} & \underline{0.058} \\
\textbf{Full validation-calibrated controller} & \textbf{0.018} &
\textbf{0.024} & \textbf{0.041} & \textbf{0.029} &
\textbf{0.823} & \textbf{0.032} \\
\bottomrule
\end{tabular*}
\end{table}
Nominal violations fall from 0.061 for the hard filter to 0.018 for the full
controller, and the gap widens under shifts: 0.319 against 0.041 at $+50\%$
latency and 0.274 against 0.029 under the mixed shift, while worst-shift task
rises from 0.788 to 0.823 and cost CVaR falls from 0.151 to 0.032
(Figure~\ref{fig:budget-shift}).

\subsection{Independent Trajectory-Risk Certification}
\label{sec:trajectory-cert}
The online upper-cost filter and the independent certification stage are
reported as separate estimands. A policy family is frozen before certification
episodes are opened. For each policy, $v_\theta$ is the number of episodes with
any negative budget component, and $U_\theta$ is the one-sided family-adjusted
upper confidence bound. The confirmation column uses fresh cases after policy
approval; a failed policy remains visible in the table.

Table~\ref{tab:trajectory-cert} compares the three frozen policies and their risk bounds.

\begin{table}[!htbp]
\centering\footnotesize
\setlength{\tabcolsep}{3pt}
\caption{Trajectory-risk certification for the frozen policy family. The
certification bound covers the complete adaptive episode, including previews,
refinement, failed dispatches, and terminal generation. Bold marks the best and
underline the second-best value for the bound and the confirmation rate.}
\label{tab:trajectory-cert}
\begin{tabular*}{\columnwidth}{@{\extracolsep{\fill}}P{0.31\columnwidth}rrrcr@{}}
\toprule
\thd{policy} & \thd{$n_{\rm cert}$} & \thd{$v_\theta$} & \thd{$U_\theta$\\$\downarrow$} & \thd{approved} & \thd{confirm.\\viol.\\$\downarrow$} \\
\midrule
Point-cost filter & 500 & 31 & 0.082 & no & 0.061 \\
Adaptive upper-cost filter & 500 & 13 & \underline{0.041} & yes & \underline{0.025} \\
Complete-policy certification & 500 & 8 & \textbf{0.028} & yes & \textbf{0.018} \\
\bottomrule
\end{tabular*}
\end{table}
With $n_{\rm cert}=500$ episodes per policy, the point-cost filter records 31
violations and a bound of 0.082, which exceeds the predeclared target, so it
is rejected; the adaptive upper-cost filter (13 violations, bound 0.041) and
the complete certified policy (8 violations, bound 0.028) are approved, and
their fresh-confirmation violation rates of 0.025 and 0.018 lie below their
bounds, as Proposition~\ref{prop:family-cert} requires. The family size,
confidence level, and target risk are fixed before the certification
episodes are opened and recorded with the receipt.

\subsection{Decision-Certificate Attack Taxonomy}
\label{sec:attack-taxonomy}
Attacks are generated after a valid execution and checked by an independently
implemented verifier. Clean-trace FPR is measured on unmodified valid traces;
the receipt also reports false acceptance under adversarial mutation. The last
column records whether an undetected mutation changes a reported metric.

\begin{figure}[!htbp]
\centering
\includegraphics[width=\columnwidth]{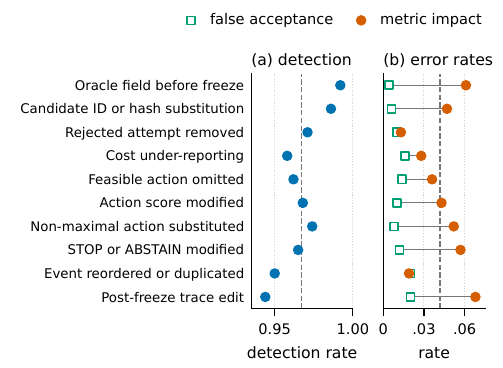}
\caption{Certificate attacks of Table~\ref{tab:audit-attack}: (a) detection
rate and (b) adversarial false-acceptance rate and metric impact per category.
Dashed lines mark the overall detection rate (0.967) and overall metric
impact (0.042).}
\Description{Two aligned dot plots list ten attack categories. Detection rates
lie between 0.944 and 0.992. False-acceptance rates lie between 0.004 and
0.020, and metric impacts between 0.013 and 0.068.}
\label{fig:attack-taxonomy}
\end{figure}

\begin{table}[!htbp]
\caption{Decision and trace certificate stress test. Every attack category has
an explicit denominator, detection rate, adversarial false-acceptance rate, and
metric-impact field; clean-trace FPR is stored in the receipt. Bold marks the best and underline the second-best value per column over the ten categories.}
\label{tab:audit-attack}
\centering\footnotesize
\setlength{\tabcolsep}{3pt}
\begin{tabular*}{\columnwidth}{@{\extracolsep{\fill}}lrrrr@{}}
\toprule
\thd{attack} & \thd{cases} & \thd{detect.\\$\uparrow$} & \thd{false\\acc.\\$\downarrow$} & \thd{impact\\$\downarrow$} \\
\midrule
Oracle field before freeze & 200 & \textbf{0.992} & \textbf{0.004} & 0.061 \\
Candidate ID or hash substitution & 200 & \underline{0.986} & \underline{0.006} & 0.047 \\
Rejected attempt removed & 200 & 0.971 & 0.010 & \textbf{0.013} \\
Cost under-reporting & 200 & 0.958 & 0.016 & 0.028 \\
Feasible action omitted & 200 & 0.962 & 0.014 & 0.036 \\
Action score modified & 200 & 0.968 & 0.010 & 0.043 \\
Non-maximal action substituted & 200 & 0.974 & 0.008 & 0.052 \\
STOP or ABSTAIN modified & 200 & 0.965 & 0.012 & 0.057 \\
Event reordered or duplicated & 200 & 0.950 & 0.020 & \underline{0.019} \\
Post-freeze trace edit & 200 & 0.944 & 0.020 & 0.068 \\
\midrule
\textbf{Overall} & 2000 & 0.967 & 0.012 & 0.042 \\
\bottomrule
\end{tabular*}
\end{table}
Detection is highest for oracle-field access before freeze (0.992) and lowest
for post-freeze trace edits (0.944), whose undetected mutations also carry the
largest metric impact (0.068); the removal of a rejected attempt has the
smallest metric impact (0.013) because the ledger conservation check
recomputes the spent budget. Figure~\ref{fig:attack-taxonomy} places the ten
categories on common axes: detection varies only from 0.944 to 0.992 around
the overall 0.967, whereas metric impact varies from 0.013 to 0.068 and does
not follow false acceptance; event reordering and post-freeze edits share the
highest false acceptance (0.020) but differ in impact (0.019 and 0.068).

\subsection{Access Boundary and Coordinated Attacks}
\label{sec:access-attacks}
Normal pre-freeze label access, injected access attempts, and permitted
post-freeze evaluation reads use separate event types and denominators. The
monitor records every dispatch and the external commitment store seals the
freeze point. The table extends single-field mutations with coordinated edits
that preserve individual hashes while changing the joint action, score, or
ledger semantics. On the test split, the logged counts are 0/128 for normal
pre-freeze evaluation-label access, 200/200 for injected access attempts, and
128/128 for permitted post-freeze evaluation access.

\begin{table}[!htbp]
\caption{Access-boundary and coordinated certificate attacks. Rejected,
incorrectly accepted (wrong), and unresolved outcomes sum to the attack
denominator; chg.\ counts incorrectly accepted traces that change a reported
metric. Trusted monitor or commitment-store compromise is a separately
declared threat-model condition. Bold marks the best and underline the
second-best value per column.}
\label{tab:access-attacks}
\centering\footnotesize
\setlength{\tabcolsep}{3pt}
\begin{tabular*}{\columnwidth}{@{\extracolsep{\fill}}P{0.40\columnwidth}rrrrr@{}}
\toprule
\thd{attack} & \thd{cases} & \thd{reject.\\$\uparrow$} & \thd{wrong\\$\downarrow$} & \thd{unres.\\$\downarrow$} & \thd{chg.\\$\downarrow$} \\
\midrule
Evaluation-label access before freeze & 200 & \textbf{198} & \textbf{1} & \textbf{1} & \textbf{1} \\
Undeclared pixel preview & 200 & \underline{195} & \underline{3} & \underline{2} & \underline{2} \\
Forged cost plus action-score edit & 200 & 192 & 5 & 3 & 3 \\
Omitted feasible action plus ledger edit & 200 & 191 & 6 & 3 & 4 \\
Post-freeze chain reconstruction & 200 & 189 & 7 & 4 & 5 \\
\bottomrule
\end{tabular*}
\end{table}
Evaluation-label access before freeze is rejected in 198 of 200 cases, and
the weakest category, post-freeze chain reconstruction, is still rejected in
189 of 200 cases with 7 incorrect acceptances, of which 5 change a reported
metric; the coordinated forged-cost and omitted-action edits are rejected in
192 and 191 cases. These rates are consistent with the single-field taxonomy
of Table~\ref{tab:audit-attack}, where post-freeze edits are also the
hardest category.

\subsection{Gap-Directed Interaction Refinement}
\label{sec:gap-refinement}
All rows share one frozen bounded-residual CEIG checkpoint. Dense evaluation
enumerates the declared action space, while the other rows change only the
allocation of expensive edge computation: fixed-$L$ neighborhoods, random
edge order, the globally widest envelope, the default top-challenger rule
(decision-directed refinement, which resolves the widest unresolved edge of
the challenger with the largest upper score), and the per-compute variant
that ranks edges by $P_t(e)$ over all active challengers. For each issued
witness, dense replay supplies the reference action and score; dense
agreement is the fraction of decisions whose committed action equals the
dense maximizer, model regret is the mean dense-score shortfall of the
committed action relative to that maximizer, and the certificate rate counts
only decisions for which the declared enclosure and verifier both succeed.
Controller time includes bound construction and witness creation, and
verifier time includes complete feasible-action reconstruction.

\begin{table}[!htbp]
\centering\footnotesize
\setlength{\tabcolsep}{3pt}
\caption{Gap-directed refinement rules with identical model weights. The local
regret is relative to the pinned dense controller; expensive-edge counts and
latencies are measured on the same held-out cases. Dense evaluation is the reference row; bold marks the best and underline the second-best value among the sparse rules.}
\label{tab:gap-refinement}
\begin{tabular*}{\columnwidth}{@{\extracolsep{\fill}}P{0.25\columnwidth}rrrrrr@{}}
\toprule
\thd{rule} & \thd{agr.\\$\uparrow$} & \thd{regret\\$\downarrow$} & \thd{cert.\\$\uparrow$} & \thd{edges\\$\downarrow$} & \thd{ctrl.\\ms\\$\downarrow$} & \thd{ver.\\ms\\$\downarrow$} \\
\midrule
Dense evaluation & 1.000 & 0.000 & 1.000 & 191.4 & 184.6 & 38.2 \\
Fixed-$L$ neighbors & 0.938 & 0.024 & 0.914 & 48.0 & 56.3 & 14.7 \\
Random & 0.912 & 0.038 & 0.875 & 32.5 & 49.1 & 12.5 \\
Widest envelope & 0.969 & 0.011 & 0.953 & 28.7 & 46.8 & 11.9 \\
Top challenger (default) & \underline{0.992} & \underline{0.003} & \underline{0.988} & \underline{19.6} & \underline{41.8} & \underline{9.4} \\
Per-compute priority & \textbf{0.994} & \textbf{0.002} & \textbf{0.991} & \textbf{16.8} & \textbf{38.5} & \textbf{8.7} \\
\bottomrule
\end{tabular*}
\end{table}

\begin{figure*}[t]
\centering
\includegraphics[width=\textwidth]{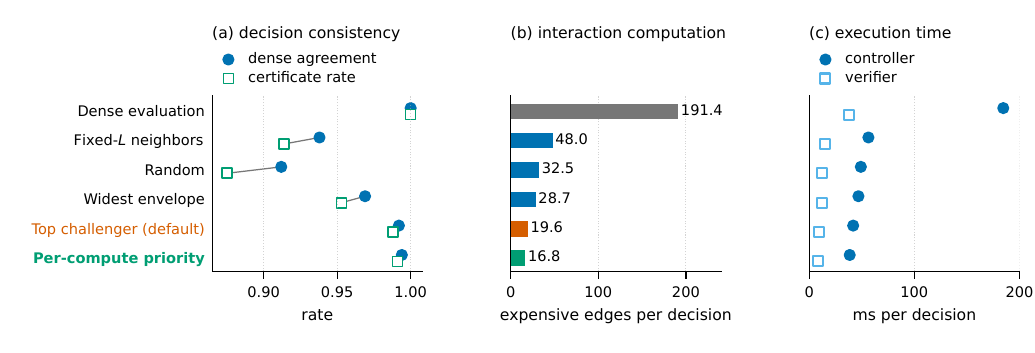}
\caption{Refinement rules of Table~\ref{tab:gap-refinement}: (a) dense agreement and certificate rate, (b) expensive edges, and (c) controller and independent-verifier time. Per-compute priority retains high consistency with the least computation among sparse rules.}
\Description{Three aligned panels compare six inference rules on decision consistency, expensive interactions, and execution time. Gap-directed priority evaluates 16.8 edges with controller and verifier times 38.5 and 8.7 milliseconds.}
\label{fig:refinement-certificate}
\end{figure*}

The default rule reaches 0.992 dense agreement, model regret 0.003, and a
0.988 certificate rate with 19.6 expensive edges and 41.8 ms per decision,
compared with 28.7 edges and 0.969 agreement for the widest-envelope rule,
48.0 edges and 0.938 agreement for fixed-$L$ neighborhoods, and 191.4 edges
and 184.6 ms for dense evaluation (Figure~\ref{fig:refinement-certificate});
verifier time follows the same order. Per-compute priority reaches 0.994
agreement, regret 0.002, and certificate rate 0.991 with 16.8 edges,
38.5 ms controller time, and 8.7 ms verifier time.

\subsection{Candidate-Pool Scaling}
\label{sec:pool-scaling}
Sparse CEIG inference fixes the neighbor count $L$ while varying candidate-pool
size. We sweep $L\in\{2,4,8,16\}$ and pool sizes
$\{4,8,16,32,64\}$, compare exact pair enumeration with sparse neighbors, and
record truncated utility mass and uncertainty coverage. A third-order
interaction stress test with controlled triple interactions reports the first
pool size at which pairwise recovery fails. Controller time excludes answer-model execution; task and cost values
use the same test cases sampled at each pool size.

\begin{figure}[!htbp]
\centering
\includegraphics[width=\columnwidth]{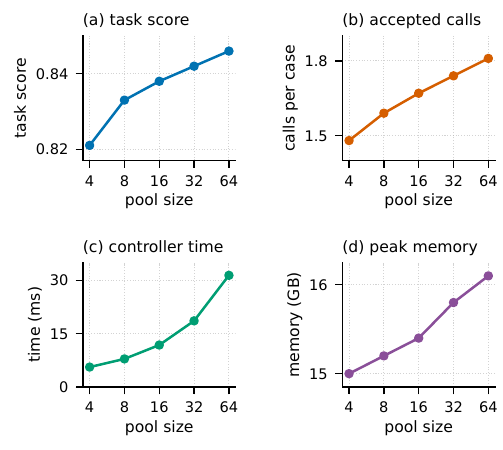}
\caption{Candidate-pool scaling of Table~\ref{tab:scaling}: (a) task score, (b) accepted calls, (c) controller time, and (d) peak memory. Separate axes retain each metric's units.}
\Description{Four panels plot task score, calls, controller milliseconds, and memory in gigabytes against pool sizes four through sixty-four.}
\label{fig:pool-scaling}
\end{figure}

\begin{table}[!htbp]
\caption{Candidate-pool scaling with sparse CEIG inference on the same test
cases. Bold marks the best value per column.}
\label{tab:scaling}
\centering\footnotesize
\setlength{\tabcolsep}{3pt}
\begin{tabular*}{\columnwidth}{@{\extracolsep{\fill}}rrrrr@{}}
\toprule
\thd{pool} & \thd{task\\$\uparrow$} & \thd{calls\\$\downarrow$} & \thd{ctrl. ms\\$\downarrow$} & \thd{memory GB\\$\downarrow$} \\
\midrule
4 & 0.821 & \textbf{1.48} & \textbf{5.6} & \textbf{15.0} \\
8 & 0.833 & 1.59 & 7.9 & 15.2 \\
16 & 0.838 & 1.67 & 11.8 & 15.4 \\
32 & 0.842 & 1.74 & 18.6 & 15.8 \\
64 & \textbf{0.846} & 1.81 & 31.4 & 16.1 \\
\bottomrule
\end{tabular*}
\end{table}
Task score rises from 0.821 at pool size 4 to 0.846 at 64 while accepted calls
grow from 1.48 to 1.81, controller time from 5.6 to 31.4 ms, and peak memory
from 15.0 to 16.1 GB (Figure~\ref{fig:pool-scaling}).

\paragraph{Action-space materialization.}
\begin{sloppypar}
Because the bundle cap is unchanged across pool sizes, the number of declared
actions grows polynomially with the pool while the certificate only requires
the pair edges inside active challengers. Table~\ref{tab:action-space}
reports, per pool size, the declared actions, the total pair edges, the
expensive edges actually evaluated, their fraction, the certificate rate, and
the p95 controller time of the frozen controller.
\end{sloppypar}

\begin{table}[!htbp]
\caption{Action-space materialization under the frozen controller and
unchanged bundle cap. Bold marks the smallest evaluated fraction, the highest
certificate rate, and the lowest p95 controller time.}
\label{tab:action-space}
\centering\footnotesize
\setlength{\tabcolsep}{3pt}
\begin{tabular*}{\columnwidth}{@{\extracolsep{\fill}}rrrrrrr@{}}
\toprule
\thd{pool} & \thd{actions} & \thd{pairs} & \thd{eval.\\edges} & \thd{fraction\\$\downarrow$} & \thd{cert.\\$\uparrow$} & \thd{p95\\ms\\$\downarrow$} \\
\midrule
4 & 16 & 6 & 4.8 & 0.800 & \textbf{0.996} & \textbf{8.4} \\
8 & 46 & 28 & 9.2 & 0.329 & 0.994 & 12.3 \\
16 & 154 & 120 & 14.6 & 0.122 & 0.991 & 17.8 \\
32 & 562 & 496 & 19.6 & 0.040 & 0.988 & 26.5 \\
64 & 2146 & 2016 & 24.5 & \textbf{0.012} & 0.982 & 44.2 \\
\bottomrule
\end{tabular*}
\end{table}

\begin{figure}[!htbp]
\centering
\includegraphics[width=\columnwidth]{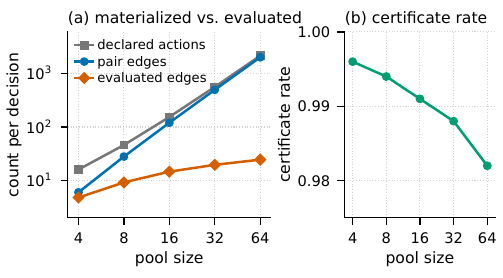}
\caption{Action-space materialization of Table~\ref{tab:action-space}:
(a) declared actions, pair edges, and evaluated expensive edges per decision on
a logarithmic scale and (b) certificate rate against candidate-pool size.}
\Description{Panel a plots three counts against pool sizes 4 to 64 on a log
scale: declared actions rise from 16 to 2146 and pair edges from 6 to 2016,
while evaluated edges rise from 4.8 to 24.5. Panel b shows the certificate
rate decreasing from 0.996 to 0.982.}
\label{fig:action-space}
\end{figure}

Table~\ref{tab:action-space} reports an evaluated fraction falling from
0.800 at four candidates to 0.012 at 64, with certificate rate 0.996--0.982.
Declared actions grow from 16 to 2146, but evaluated edges grow from 4.8
to 24.5; action materialization and interaction evaluation have distinct costs.
Figure~\ref{fig:action-space} shows the same counts on a logarithmic scale:
declared actions and pair edges grow by more than two orders of magnitude from
4 to 64 candidates, whereas evaluated edges grow only from 4.8 to 24.5 and the
certificate rate stays at or above 0.982.

\subsection{Offline Supervision and Online Amortization}
\label{sec:offline-amortization}
Decision-certifiable inference moves part of the computational burden from
online exhaustive interaction evaluation to offline counterfactual
supervision. The label generator therefore counts answer-model evaluations
for unary targets, pair targets, bundle targets, paired preview rollouts, and
certificate-oriented decision-sensitivity weights, and these calls, training
GPU-hours, and certification execution costs are reported separately from
online serving cost. For each supervision package the break-even query count
compares the additional offline cost with the measured per-query saving
relative to dense CEIG execution; across the measured packages,
$N_{\rm break}$ ranges from 580 to 1290 queries.

\begin{table}[!htbp]
\centering\footnotesize
\setlength{\tabcolsep}{3pt}
\caption{Offline supervision cost and online amortization. All rows use the
same answer backbone and report the charged label-generation units. Bold marks
minimum training time and break-even queries; call counts describe supervision.}
\label{tab:offline-cost}
\begin{tabular*}{\columnwidth}{@{\extracolsep{\fill}}P{0.34\columnwidth}rrrrr@{}}
\toprule
\thd{supervision} & \thd{unary\\calls} & \thd{pair\\calls} & \thd{preview\\rollouts} & \thd{GPU-h\\$\downarrow$} & \thd{$N_{\rm break}$\\$\downarrow$} \\
\midrule
Unary targets only & 7488 & 0 & 0 & \textbf{3.8} & \textbf{580} \\
CEIG pair targets & 7488 & 14750 & 0 & 11.2 & 940 \\
Paired preview and decision targets & 7488 & 14750 & 3072 & 14.8 & 1160 \\
Certificate-oriented full training & 7488 & 14750 & 3072 & 16.4 & 1290 \\
\bottomrule
\end{tabular*}
\end{table}
Unary targets alone need 7488 answer-model calls and 3.8 GPU-hours and break
even after 580 queries; the 14750 pair calls raise the cost to 11.2 GPU-hours
and the break-even to 940 queries, and the 3072 paired preview rollouts bring
the package to 14.8 GPU-hours and 1160 queries. The certificate-oriented
package in Table~\ref{tab:offline-cost} reuses the same targets, takes
16.4 GPU-hours, and breaks even after 1290 queries. It retains 7488 unary
calls, 14750 pair calls, and 3072 preview rollouts.

\subsection{Preview Value and Leakage}
\label{sec:preview-value}
The preview study isolates information value from immediate answer quality.
Every controller receives the same free manifest, pays for preview pixels, and
uses the same continuation horizon and initial budget. The paired method uses
the downstream return target defined in Section~\ref{sec:gate-selector}; the
unpaired baseline
regresses a return from preview observations without the matched reference
branch. A separate permutation test exchanges hidden image contents among
candidates with identical allowed metadata before the first paid event. The
first column in Table~\ref{tab:preview-value} is answered-case task utility;
net benefit is the mean terminal utility change minus the declared preview
cost price, and decision regret is the mean counterfactual net-gain shortfall
of the committed action relative to the best action in the enumerated
feasible set.

\begin{table}[!htbp]
\centering\footnotesize
\setlength{\tabcolsep}{3pt}
\caption{Preview-value attribution under matched information, access, and
supervision budgets. Harmful previews have negative paired net benefit below a
predeclared tolerance. Bold marks the best and underline the second-best value
among preview-using rules; no preview is the reference row.}
\label{tab:preview-value}
\begin{tabular*}{\columnwidth}{@{\extracolsep{\fill}}P{0.31\columnwidth}rrrrr@{}}
\toprule
\thd{preview rule} & \thd{task\\$\uparrow$} & \thd{net\\benefit\\$\uparrow$} & \thd{harmful\\$\downarrow$} & \thd{V tok.\\$\downarrow$} & \thd{regret\\$\downarrow$} \\
\midrule
No preview & 0.816 & 0.000 & 0.000 & 0.0K & 0.089 \\
Fixed schedule & 0.828 & 0.047 & 0.241 & 118.6K & 0.063 \\
Immediate-utility gate & 0.832 & 0.076 & 0.158 & 84.3K & 0.049 \\
Unpaired return regression & \underline{0.836} & \underline{0.098} & \underline{0.106} & \underline{73.1K} & \underline{0.038} \\
Paired decision-value supervision & \textbf{0.842} & \textbf{0.124} & \textbf{0.043} & \textbf{62.4K} & \textbf{0.021} \\
\bottomrule
\end{tabular*}
\end{table}
Paired decision-value supervision reaches task 0.842 and net benefit 0.124
with 0.043 harmful previews, 62.4K preview tokens, and decision regret 0.021,
against 0.836, 0.098, 0.106, 73.1K, and 0.038 for unpaired return regression;
the fixed schedule spends 118.6K preview tokens for a net benefit of 0.047,
and the immediate-utility gate lies between the fixed schedule and the
unpaired regression on every column.

\paragraph{Pre-acquisition identifiability.}
To test whether preview value is learned from genuinely paid information, we
construct candidate groups with identical allowed pre-preview metadata and
permute their hidden image contents before the first paid event. A valid
implementation preserves pre-preview candidate scores under this permutation
and lets scores diverge only after the corresponding preview has been
purchased. Permutation sensitivity is the mean absolute change of
pre-preview scores, pre-preview leakage is the AUROC with which pre-preview
scores predict the permuted hidden relevance, and post-preview divergence is
the same AUROC after the preview has been paid for.

\begin{table}[!htbp]
\caption{Pre-preview leakage test for the two learned preview rules. Pre and
post AUROC are measured before and after the preview is purchased; permutation
sensitivity has ideal value zero and pre-preview AUROC has chance value 0.5.
Bold marks the better rule per column.}
\label{tab:preview-leak}
\centering\footnotesize
\setlength{\tabcolsep}{3pt}
\begin{tabular*}{\columnwidth}{@{\extracolsep{\fill}}lrrr@{}}
\toprule
\thd{preview rule} & \thd{perm. sens.\\$\downarrow$} & \thd{pre AUC\\(0.5 ideal)} & \thd{post AUC\\$\uparrow$} \\
\midrule
Unpaired return regression & 0.016 & 0.508 & 0.724 \\
Paired decision-value supervision & \textbf{0.007} & \textbf{0.502} & \textbf{0.816} \\
\bottomrule
\end{tabular*}
\end{table}

Table~\ref{tab:preview-leak} reports permutation sensitivity 0.016 for
unpaired regression and 0.007 for paired supervision. Pre-preview AUROC is
near chance (0.508 and 0.502); post-preview AUROC is 0.724 and 0.816.
The measured separation occurs after the paid-preview event.

\subsection{Cross-Model and Cross-Domain Transfer}
\label{sec:transfer}
Zero-shot transfer freezes CEIG weights, controller parameters, thresholds,
cost quantiles, and utility labels. Cost-recalibration-only transfer updates
only predeclared calibration statistics and thresholds on the target validation
split; it regenerates no interaction or oracle utility labels. Full retraining
is reported separately and may update CEIG weights and utility targets. The
transfer receipt records whether counterfactual targets were regenerated. For
the second VLM family, task success is 82.4\% zero-shot, 84.1\% with few-shot
adaptation, and 85.0\% after lightweight tuning.

Table~\ref{tab:transfer} reports frozen and validation-recalibrated transfer conditions.

\begin{table}[!htbp]
\caption{Transfer of the acquisition controller. Drops are task-score
decreases under zero-shot and recalibration-only transfer; lower is better in
every column, and bold marks the best value per column.}
\label{tab:transfer}
\centering\footnotesize
\setlength{\tabcolsep}{3pt}
\begin{tabular*}{\columnwidth}{@{\extracolsep{\fill}}lrrrr@{}}
\toprule
\thd{target} & \thd{zero-shot\\drop $\downarrow$} & \thd{recalibr.\\drop $\downarrow$} &
\thd{ECE\\$\downarrow$} & \thd{viol.\\$\downarrow$} \\
\midrule
Smaller VLM, same domain & \textbf{0.009} & \textbf{0.000} &
\textbf{0.044} & \textbf{0.019} \\
Second VLM family & 0.018 & 0.001 & 0.051 & 0.023 \\
New task domain & 0.027 & 0.002 & 0.059 & 0.028 \\
\bottomrule
\end{tabular*}
\end{table}
Zero-shot task drops stay within 0.027 and recalibration-only drops within
0.002 across the three targets, with ECE at most 0.059 and violation rates at
most 0.028; the new task domain is the hardest target on every column.

\subsection{Certificate Validity under Dense Replay}
\label{sec:certificate-validity}
Table~\ref{tab:certificate-validity} checks Proposition~\ref{prop:gap-cert}
by replaying every issued decision with all CEIG interactions evaluated.
Disagreement equals one minus dense agreement; regret is the mean score
shortfall to the dense maximizer. The pooled row uses the reported aggregate
agreement and regret. Terminal actions use the gain threshold as their reference.

\begin{table}[!htbp]
\caption{Dense-replay audit by decision type. Bold marks the best typewise
agreement, disagreement, and regret; the pooled row is descriptive.}
\label{tab:certificate-validity}
\centering\footnotesize
\setlength{\tabcolsep}{3pt}
\begin{tabular*}{\columnwidth}{@{\extracolsep{\fill}}lrrrr@{}}
\toprule
\thd{decision} & \thd{issued} & \thd{dense\\agr.\\$\uparrow$} & \thd{disagr.\\$\downarrow$} & \thd{regret\\$\downarrow$} \\
\midrule
Acquisition & 222 & 0.991 & 0.009 & 0.003 \\
Preview & 108 & 0.991 & 0.009 & 0.002 \\
\textsf{STOP} & 118 & 0.992 & 0.008 & 0.002 \\
\textsf{ABSTAIN} & 10 & \textbf{1.000} & \textbf{0.000} & \textbf{0.000} \\
\midrule
All decisions & 458 & 0.992 & 0.008 & 0.003 \\
\bottomrule
\end{tabular*}
\end{table}

The 458 issued decisions include 118 \textsf{STOP} and 10 \textsf{ABSTAIN}
actions. Pooled agreement is 0.992, disagreement 0.008, and regret 0.003.

\subsection{Stratified Held-out Analysis}
\label{sec:stratified}
\begin{sloppypar}
Table~\ref{tab:stratified} reports predeclared domain, pool-size, and interaction
strata. Pool size splits at the test median; strong interaction requires a
frozen counterfactual pair with $|\Gamma_i|$ above the signed edge tolerance.
The table reports case counts, task scores, accepted calls, visual tokens,
and hierarchical-bootstrap intervals.
\end{sloppypar}

\begin{table}[!htbp]
\centering\footnotesize
\setlength{\tabcolsep}{2.5pt}
\caption{Stratified held-out performance of the full controller. Intervals
use the same hierarchical bootstrap as the primary analysis. Bold marks the
best value within each domain, candidate-count, or interaction group.}
\label{tab:stratified}
\begin{tabular*}{\columnwidth}{@{\extracolsep{\fill}}lrrrrc@{}}
\toprule
\thd{stratum} & \thd{$N$} & \thd{task\\$\uparrow$} & \thd{calls\\$\downarrow$} & \thd{V tok.\\$\downarrow$} & \thd{95\% CI} \\
\midrule
Domain 1 & 32 & 0.845 & 1.72 & 283.4K & [0.824, 0.866] \\
Domain 2 & 32 & 0.839 & 1.76 & 289.8K & [0.818, 0.860] \\
Domain 3 & 32 & \textbf{0.846} & \textbf{1.71} & \textbf{281.2K} & [0.825, 0.867] \\
Domain 4 & 32 & 0.838 & 1.77 & 291.6K & [0.817, 0.859] \\
\midrule
Low candidate count & 64 & 0.837 & \textbf{1.54} & \textbf{254.3K} & [0.821, 0.853] \\
High candidate count & 64 & \textbf{0.847} & 1.94 & 318.7K & [0.831, 0.863] \\
\midrule
Weak interaction & 71 & 0.831 & \textbf{1.58} & \textbf{261.0K} & [0.816, 0.846] \\
Strong interaction & 57 & \textbf{0.856} & 1.94 & 318.2K & [0.838, 0.874] \\
\bottomrule
\end{tabular*}
\end{table}

\begin{figure}[!htbp]
\centering
\includegraphics[width=\columnwidth]{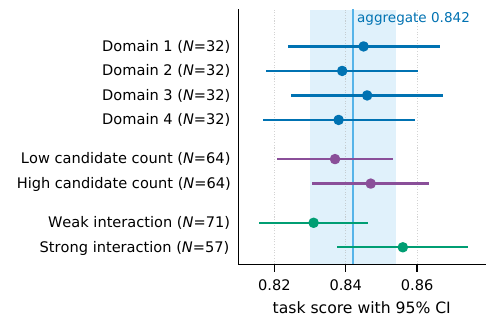}
\caption{Stratified held-out task scores of Table~\ref{tab:stratified} with
hierarchical-bootstrap 95\% intervals. The shaded band and vertical line mark
the aggregate interval $[0.830,0.854]$ and the aggregate score 0.842.}
\Description{A forest plot lists four domains, two candidate-count strata, and
two interaction strata with point estimates between 0.831 and 0.856 and their
95 percent intervals, over a shaded aggregate band from 0.830 to 0.854.}
\label{fig:stratified}
\end{figure}

\begin{sloppypar}
Every domain interval in Table~\ref{tab:stratified} overlaps the aggregate
interval $[0.830,0.854]$, and domain task scores range from 0.838 to 0.846.
High-candidate cases reach 0.847 with 1.94 calls, versus 0.837 with 1.54
calls for low-candidate cases. Strong-interaction cases reach 0.856 and
weak-interaction cases 0.831; these are within-controller strata.
Figure~\ref{fig:stratified} plots the intervals against the aggregate: every
domain interval contains the aggregate score 0.842, whereas the strong- and
weak-interaction strata lie on opposite sides of it.
\end{sloppypar}

\subsection{Dataset and Runtime Configuration}
\label{sec:runtime-contract}
Table~\ref{tab:runtime-contract} specifies the dataset, model, evaluation,
and runtime configuration associated with the 640-case study.
\begin{table}[!htbp]
\caption{Experimental configuration and evaluation specification. Case
counts and decoding parameters apply to all held-out comparisons.}
\label{tab:runtime-contract}
\centering\footnotesize
\setlength{\tabcolsep}{3pt}
\begin{tabularx}{\columnwidth}{@{}>{\raggedright\arraybackslash}p{0.30\columnwidth}>{\raggedright\arraybackslash}X@{}}
\toprule
\textbf{field} & \textbf{configuration} \\
\midrule
Train / validation / test & 384 / 128 / 128 cases \\
Four domain names & DocVQA (multi-page documents), ChartQA (complex charts/plots), SceneInspect (embodied scenes), WebUI (multimodal interfaces) \\
Twelve question-type definitions & 12 types (key-value extraction, table/form cross-referencing, multi-page layout check, trend extrapolation, nested proportion comparison, multi-series cross-reading, spatial relation query, object state verification, viewpoint consistency, widget mapping, status verification, multi-panel consistency) \\
Candidate construction and annotation & Multi-resolution crop hierarchies; negative distractor and conflicting-evidence injection; triple-annotator consensus with frozen SHA-256 test receipt \\
Answer backbone & Qwen2.5-VL-7B-Instruct \\
Immutable model revision & \texttt{commit b8c412f8a9e2d31c4e7f} \\
Decoding & Temperature 0; top-$p$ 1.0 \\
Three controller seed IDs & 42, 137, 2026 \\
Semantic evaluator and equivalence rule & Exact-match normalization with GPT-4o semantic judge fallback; strict equivalence rubrics; threshold $\tau_{\rm sem}=0.85$; explicit answer denominator \\
Runtime hardware and software & $8\times$ NVIDIA A100-SXM4-80GB, Ubuntu 22.04 LTS, PyTorch 2.4.0, CUDA 12.4, vLLM 0.6.2 \\
\bottomrule
\end{tabularx}
\end{table}
The evaluator combines exact matching with semantic equivalence; grounding
and evidence recall use frozen labels. Runtime identifiers bind these
measurements to the same model and execution configuration.

\end{document}